%% file: acl_latex.tex
\pdfoutput=1

\documentclass[11pt]{article}

\usepackage[preprint]{acl}
\usepackage{pifont}
\usepackage{times}
\usepackage{latexsym}

\usepackage{booktabs}

\usepackage{paralist}

\usepackage[T1]{fontenc}

\usepackage[utf8]{inputenc}

\usepackage{microtype}

\usepackage{inconsolata}
\usepackage{bbding}
\usepackage{graphicx}
\usepackage{subcaption}
\usepackage{mdframed}
\usepackage{letltxmacro}  

\newcommand\blfootnote[1]{%
  \begingroup
  \renewcommand\thefootnote{}\footnote{#1}%
  \addtocounter{footnote}{-1}%
  \endgroup
}

\newcommand{\mySymbolScale}{0.7}

\LetLtxMacro{\originalCheckmark}{\Checkmark} 
\renewcommand{\Checkmark}{\scalebox{\mySymbolScale}{\originalCheckmark}} 

\LetLtxMacro{\originalXSolid}{\XSolid} 
\renewcommand{\XSolid}{\scalebox{\mySymbolScale}{\originalXSolid}} 

\title{WangchanThaiInstruct: An instruction-following Dataset for Culture-Aware, Multitask, and Multi-domain Evaluation in Thai}

\author{
Peerat Limkonchotiwat\textsuperscript{$\spadesuit$*}, 
Pume Tuchinda\textsuperscript{$\heartsuit$*},
Lalita Lowphansirikul\textsuperscript{$\heartsuit$}, \\
\textbf{Surapon Nonesung}\textsuperscript{$\heartsuit$}, 
\textbf{Panuthep Tasawong}\textsuperscript{$\heartsuit$}, 
\textbf{Alham Fikri Aji}\textsuperscript{$\diamondsuit$}, \\
\textbf{Can Udomcharoenchaikit}\textsuperscript{$\heartsuit$},
\textbf{Sarana Nutanong}\textsuperscript{$\heartsuit$}\\
  \textsuperscript{$\spadesuit$}AI Singapore, 
  \textsuperscript{$\heartsuit$}VISTEC, 
  \textsuperscript{$\diamondsuit$}MBZUAI \\
  \texttt{peerat@aisingapore.org}, \texttt{pumet\_pro@vistec.ac.th}
  }

\begin{document}
\maketitle
\blfootnote{\textsuperscript{*}Equal contributions}
\begin{abstract}

Large language models excel at instruction-following in English, but their performance in low-resource languages like Thai remains underexplored. 
Existing benchmarks often rely on translations, missing cultural and domain-specific nuances needed for real-world use. 
We present ThaiInstruct, a human-authored Thai dataset for evaluation and instruction tuning, covering four professional domains and seven task types. 
Created through a multi-stage quality control process with annotators, domain experts, and AI researchers, ThaiInstruct supports two studies: (1) a zero-shot evaluation showing performance gaps on culturally and professionally specific tasks, and (2) an instruction tuning study with ablations isolating the effect of native supervision. 
Models fine-tuned on ThaiInstruct outperform those using translated data in both in-domain and out-of-domain benchmarks.
These findings underscore the need for culturally and professionally grounded instruction data to improve LLM alignment in low-resource, linguistically diverse settings.\footnote{Dataset and Model Collection:\url{https://huggingface.co/collections/airesearch/wangchan-thai-instruction-6835722a30b98e01598984fd}}

\end{abstract}

\input{01_introduction}

\input{02_related_works}
\input{03_wangchan_instruct}

\input{04_experimental_setup}
\input{05_results}

\input{06_analysis}

\input{07_conclusion}

\subsection*{Acknowledgement}

This research is supported by the National Research Foundation, Singapore, under its National Large Language Models Funding Initiative. Any opinions, findings, and conclusions or recommendations expressed in this material are those of the author(s) and do not reflect the views of the National Research Foundation, Singapore.

\section*{Limitations}
While our dataset and benchmark are a step forward toward better representations of Thai in the LLM era, we acknowledge
that significant progress is still to be made. We
outline several limitations of our study below.

\noindent
\textbf{Evaluate Metric}. Since the main metric of our benchmark relies on LLM-as-a-judge, reproducing this number in the future might not be straightforward since the API model might update.
However, using traditional metrics like BLEU or ROUGE-L also yields the worst evaluation metric, as we discussed in Appendix~\ref{subsec:metric_human}.

\noindent
\textbf{Judge Models}. As we discussed in Appendix~\ref{subsec:metric_human}, the limitation of our evaluation metric is the judge model.
We found that there are no dominant models to judge Thai knowledge and culture, where the Kendall between annotators and LLM is only 0.1259 points.

\noindent
\textbf{Non-Commercial Use Limitation}. A significant portion of our released dataset (30,000 samples) is distributed under the CC BY-NC license. This restricts the use of these samples to non-commercial purposes only. While this allows for academic and research use, it can be limiting for potential users who require the data for commercial applications. The remaining samples (5,014 samples) under the more permissive CC BY-SA 4.0 license can be used commercially under the terms of that license.

\section*{Ethical Statement}
We hired all annotators from an annotation company in Thailand. However, for domain experts, we directly contacted their companies and compensated them at a rate higher than Thailand's minimum wage. Similarly, our AI researchers received the same payment rate as the domain experts.
Regarding data licensing, we plan to publicly release the datasets under a CC-BY-SA and NC license. This choice accommodates certain data sources that, despite permitting collection, requested a NonCommercial (NC) license.

\bibliography{custom}

\clearpage
\onecolumn
\appendix

\label{sec:appendix}
\input{appendix/appendix}

\end{document}

%% file: 01_introduction.tex
\section{Introduction}

Large language models (LLMs) are crucial in NLP applications due to their instruction following capabilities, enabling zero-shot/few-shot learning for diverse tasks (e.g., summarization, machine translation) and eliminating task-specific model training.
However, evaluation of LLMs remains centered on English, such as Open LLM Leaderboard~\cite{mtbench} and MT-Bench~\cite{openllmleaderboard}, leaving performance in underrepresented languages like Thai under-investigated.
This gap is especially critical in domain-specific applications---\emph{such as legal, financial, medical, and retail use cases}---where professional knowledge and cultural grounding are required.

Recent efforts have extended LLM evaluations to Southeast Asian languages---including Indonesian, Vietnamese, and Thai---through benchmarks like SEA-Bench~\cite{seabench}, SEA-HELM~\cite{seahelm}, and SEA-Crowd~\cite{seacrowd}. 
However, these benchmarks predominantly rely on translated English data and lack domain specificity. 
This gap risk inflated performance estimates that do not reflect actual usage in native, domain-sensitive contexts.
In the specific case of Thai, the problem is further exacerbated by the lack of accessible, high-quality native-language benchmarks. 
While authentic resources such as ThaiH6 and ThaiCLI~\cite{thai-h6-cli} have been developed, they remain closed-source and unavailable to the broader research community, hindering transparent and reproducible evaluation of LLM performance on native Thai instructions.

\begin{figure*}[h!]
  \centering
  \includegraphics[width=\textwidth]{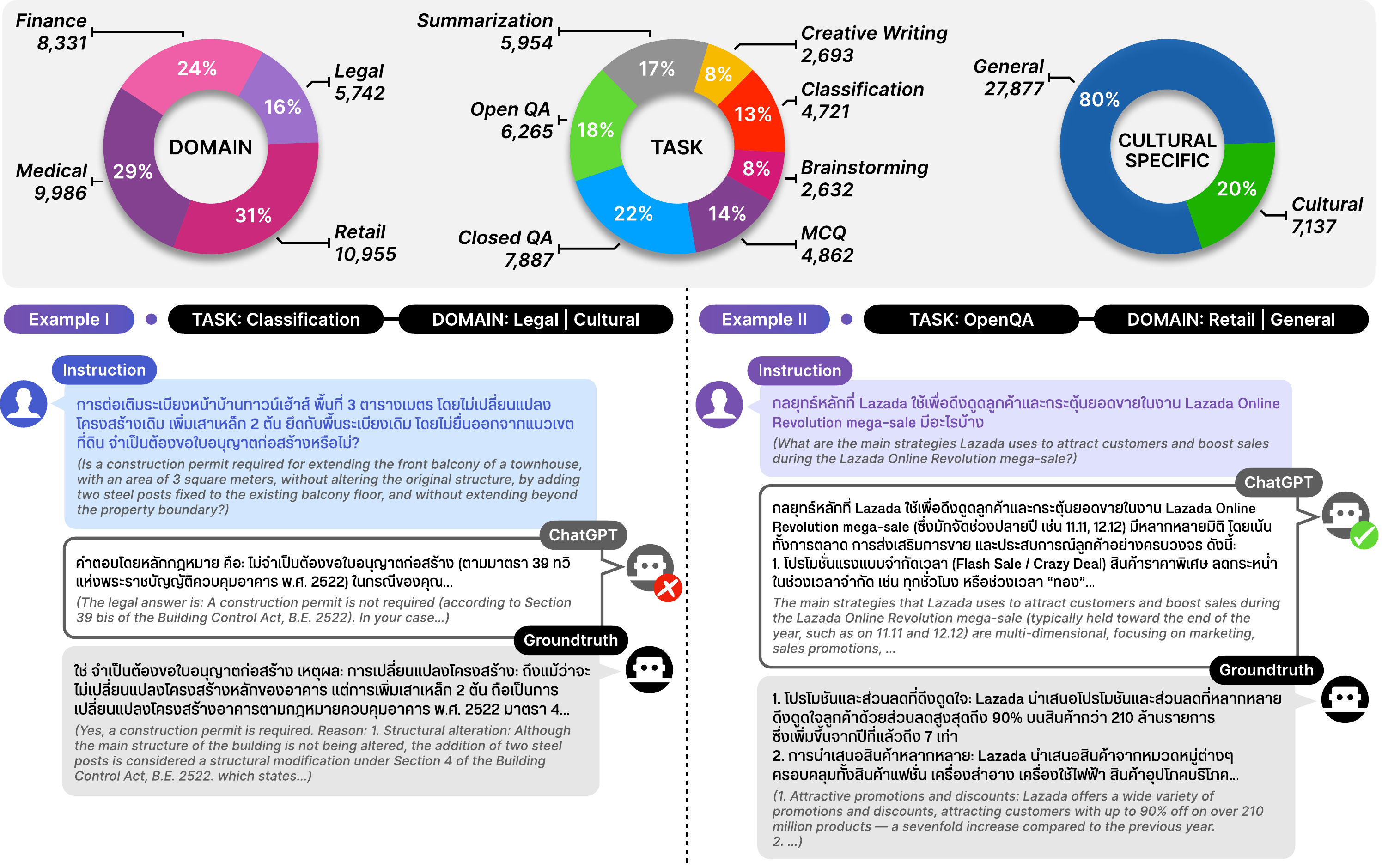}
  \vspace{-8mm}
    \caption{
    \textbf{(Top)} Sample distribution by domain and task type, including culture-specific vs. general data.
\textbf{(Bottom)} Examples from WangchanThaiInstruct with a sample ChatGPT-4o responses. (\textbf{Left)} A culturally specific question on Thai construction law, where ChatGPT contradicts the ground truth about permit requirements for balcony extensions. \textbf{(Right)} A general question on Lazada's marketing strategies, correctly answered by ChatGPT.
    }
    \vspace{-5mm}
  \label{fig:figure1}
\end{figure*}

%
%

Moreover, developing LLMs for Thai remains challenging due to the lack of high-quality supervised datasets.
Although recent advances such as Llama~\cite{llama3}, Gemma~\cite{gemma2}, SEA-LION~\cite{sealion}, and Qwen~\cite{qwen25} have demonstrated promising support for Thai,  their understanding of culturally and professionally grounded instructions remains underexplored.
%
As shown in Figure~\ref{fig:figure1}, state-of-the-art models like ChatGPT-4o struggle to respond accurately to instructions rooted in Thai cultural and domain-specific contexts, despite their general fluency.
Motivated by these challenges, we aim to address two research questions: 
\textbf{RQ1}: \emph{``How effectively do current large language models (LLMs) handle Thai culturally and professionally specific instructions?''}
%
%
\textbf{RQ2}: \emph{``To what extent does fine-tuning large language models on native Thai data improve their culturally and professionally specific understanding and response accuracy?''}


To address these RQs, we present WangchanThaiInstruct, a Thai instruction-following dataset for evaluating and improving LLM performance on Thai instructions.
WangchanThaiInstruct comprises 28,098 training and 6,916 test samples across four domains (\emph{Medical, Law, Finance, and Retail}) and seven task types (\emph{Brainstorming, Classification, Closed QA, Creative writing, Multiple choice, Open QA, and Summarization}).
All samples are human-authored, with no reliance on LLM-generated content.
We also introduce an evaluation protocol aligned with human preferences, using an LLM-as-a-judge approach to reflect Thai cultural and domain-specific contexts.

%

%
%
%

Using WangchanThaiInstruct, we conducted two studies.
For \textbf{RQ1}, we evaluated state-of-the-art LLMs in a zero-shot setting with the test split, and an LLM-as-a-judge protocol to assess factual accuracy and reasoning. 
We analyzed model performance across culturally and professionally grounded instructions.
For \textbf{RQ2}, we fine-tuned LLMs on the training split and compared them to models trained on translated datasets like Alpaca and Dolly. 
Ablation studies isolate the effect of native supervision under matched data size and format, revealing how domain-specific data improves alignment.


Experimental results show that zero-shot LLMs struggle with culturally and domain-specific Thai instructions, especially in legal and multiple-choice tasks. 
Reasoning evaluations reveal poor rationale quality, with models often failing our judgment-based metric despite fluent outputs.
%
In contrast, WangchanThaiInstruct-tuned models show clear gains in both in-domain and out-of-domain settings. 
These findings highlight the need for deeper alignment with linguistic, cultural, and professional norms in Thai. 

\begin{table*}[h!]
\centering
\renewcommand{\arraystretch}{1}
\resizebox{\textwidth}{!}{%
\begin{tabular}{lcccccccc}
\hline
\multicolumn{1}{l|}{\textbf{Dataset}} &
  \multicolumn{1}{c|}{\textbf{\#Tasks}} &
  \multicolumn{1}{c|}{\textbf{\#Domains}} &
  \multicolumn{1}{c|}{\textbf{Cultural Data?}} &
  \multicolumn{1}{c|}{\textbf{Training\&Test Data?}} &
  \multicolumn{1}{c|}{\textbf{Human QC?}} &
  \multicolumn{1}{c|}{\textbf{Expert QC?}} &
  \multicolumn{1}{c|}{\textbf{Human Craft Data?}} &
  \textbf{License} \\ \hline
\multicolumn{9}{c}{\textit{\textbf{Multilingual Benchmarks}}} \\ \hline
\multicolumn{1}{l|}{SEACrowd} &
  \multicolumn{1}{c|}{2} &
  \multicolumn{1}{c|}{2} &
  \multicolumn{1}{c|}{\XSolid} &
  \multicolumn{1}{c|}{\XSolid} &
  \multicolumn{1}{c|}{\XSolid\,/\,\Checkmark} &
  \multicolumn{1}{c|}{\XSolid} &
  \multicolumn{1}{c|}{\XSolid\,/\,\Checkmark} &
  Mixed \\ 
\multicolumn{1}{l|}{SEA-Bench} &
  \multicolumn{1}{c|}{1} &
  \multicolumn{1}{c|}{1} &
  \multicolumn{1}{c|}{\XSolid} &
  \multicolumn{1}{c|}{\XSolid} &
  \multicolumn{1}{c|}{\Checkmark} &
  \multicolumn{1}{c|}{\XSolid} &
  \multicolumn{1}{c|}{\Checkmark} &
  Apache-2.0 \\ 
\multicolumn{1}{l|}{SEA-HELM} &
  \multicolumn{1}{c|}{4} &
  \multicolumn{1}{c|}{2} &
  \multicolumn{1}{c|}{\XSolid} &
  \multicolumn{1}{c|}{\XSolid} &
  \multicolumn{1}{c|}{\XSolid\,/\,\Checkmark} &
  \multicolumn{1}{c|}{\XSolid} &
  \multicolumn{1}{c|}{\XSolid\,/\,\Checkmark} &
  CC BY-SA \\ \hline
\multicolumn{9}{c}{\textit{\textbf{Thai Benchmarks}}} \\ \hline
\multicolumn{1}{l|}{ThaiLLM Leaderboard} &
  \multicolumn{1}{c|}{3} &
  \multicolumn{1}{c|}{2} &
  \multicolumn{1}{c|}{\XSolid} &
  \multicolumn{1}{c|}{\XSolid} &
  \multicolumn{1}{c|}{\XSolid\,/\,\Checkmark} &
  \multicolumn{1}{c|}{\XSolid} &
  \multicolumn{1}{c|}{\XSolid\,/\,\Checkmark} &
  MIT \\ 
\multicolumn{1}{l|}{Thai-H6} &
  \multicolumn{1}{c|}{1} &
  \multicolumn{1}{c|}{1} &
  \multicolumn{1}{c|}{\XSolid} &
  \multicolumn{1}{c|}{\XSolid} &
  \multicolumn{1}{c|}{\Checkmark} &
  \multicolumn{1}{c|}{\XSolid} &
  \multicolumn{1}{c|}{\XSolid} &
  Closed-source \\ 
\multicolumn{1}{l|}{ThaiCLI} &
  \multicolumn{1}{c|}{1} &
  \multicolumn{1}{c|}{1} &
  \multicolumn{1}{c|}{\Checkmark} &
  \multicolumn{1}{c|}{\XSolid} &
  \multicolumn{1}{c|}{\Checkmark} &
  \multicolumn{1}{c|}{\Checkmark} &
  \multicolumn{1}{c|}{\Checkmark} &
  Closed-source \\ 
\multicolumn{1}{l|}{\emph{WangchanThaiInstruct (Ours)}} &
  \multicolumn{1}{c|}{7} &
  \multicolumn{1}{c|}{4} &
  \multicolumn{1}{c|}{\Checkmark} &
  \multicolumn{1}{c|}{\Checkmark} &
  \multicolumn{1}{c|}{\Checkmark} &
  \multicolumn{1}{c|}{\Checkmark} &
  \multicolumn{1}{c|}{\Checkmark} &
  CC BY-SA/NC \\ \hline \hline
\end{tabular}%
}
\vspace{-3mm}
\caption{Benchmark comparison between previous works and ours. For multilingual benchmarks, the number is displayed only for Thai data in each benchmark. Note that when we put both \XSolid\,and\,\Checkmark, meaning that the benchmark combines both choices, depending on the samples.}
\vspace{-5mm}
\label{tab:data_statistic}
\end{table*}

The contributions of our paper are as follows:
\begin{compactitem}[\hspace{3mm}•]
\item We introduce WangchanThaiInstruct, a human-authored Thai instruction dataset spanning multiple domains and task types, designed to support both cultural evaluation and instruction tuning in real-world, context-sensitive applications.

\item We conduct two structured studies: a zero-shot evaluation revealing performance gaps on culturally and professionally specific Thai instructions, and an instruction tuning study with ablation experiments offering practical guidance on how to use WangchanThaiInstruct effectively.

\item We present a reproducible dataset development process, covering sourcing, task design, cultural annotation, and multi-stage quality control, generalizable to other underrepresented languages.
\underline{We release all artifacts}, including the dataset, evaluation splits, training scripts, and all fine-tuned models, including those with SOTA results on Thai LLM benchmarks, to provide strong baselines, promote reproducibility, and enable future research on applications that require Thai-specific knowledge.

\end{compactitem}





%% file: 02_related_works.tex
\section{Related Works}

\subsection{Thai Language Models}
The development of Thai language models has gained increasing momentum in recent years.
WangchanBERTa~\cite{wangchanberta} represented one of the first pre-trained Thai language models based on BERT~\cite{bert}. 
Followed by PhayaThaiBERT~\cite{phayathaibert}, which expanded on this approach through expanding its tokenizer with foreign vocabularies to better handle loaned words.
Following the release of GPT~\cite{gpt3} style models, multiple works adapted these architectures for Thai. 
Recently, works like OpenThaiGPT~\cite{openthaigpt15} made significant progress by finetuning base foundation LLMs like Qwen or Llama on large synthetic Thai instruction tuning datasets.
Meanwhile, Typhoon~\cite{typhoon2} represents the state-of-the-art in Thai LLMs, extending its support to multimodal domains like speech and vision. 
%
%
 While progress has been made, many works heavily rely on machine-translated instruction datasets often lacking Thai nuances, idiomatic expressions, and cultural context. This underscores the need for high-quality, culturally grounded, human-curated datasets reflecting real-world usage.

\subsection{Benchmark for Thai Language Models}
Recently, there has been growing interest in evaluating LLMs on Southeast Asian (SEA) data.
As shown in Table~\ref{tab:data_statistic}, efforts such as SEACrowd \cite{seacrowd} and SEA-Helm \cite{seahelm} extend classical LLM benchmarks to Southeast Asian languages by collecting and verifying machine-translated samples through large-scale crowd sourcing.
However, these resources often rely on generic questions that do not adequately probe deeper cultural nuances of the target language.
Subsequent work like SEA-Bench \cite{seabench} incorporates regional exams to introduce greater contextual relevance, yet still falls short of reflecting the diverse, real-world scenarios encountered by native speakers.
Within Thai specifically, efforts like the ThaiLLM Leaderboard\footnote{\url{https://huggingface.co/spaces/ThaiLLM-Leaderboard/leaderboard}} curated Thai-specific evaluation datasets that still heavily rely on machine-translated data and a small-scale set of manually written MT-Bench\footnote{\url{https://huggingface.co/datasets/ThaiLLM-Leaderboard/mt-bench-thai}} style evaluation, limiting their effectiveness for comprehensive evaluation.
Later efforts, such as Thai-H6 and ThaiCLI \cite{thai-h6-cli}, introduced benchmarks aimed at assessing cultural understanding in Thai.
Nonetheless, these remain closed-source and lack accompanying training datasets, limiting their accessibility to the broader research community.

%% file: 03_wangchan_instruct.tex
\section{Dataset and Study Design}

As mentioned in the introduction section, we aim to address two core research questions.
%
%
To answer these questions, we design two studies using WangchanThaiInstruct: (1) a zero-shot evaluation to assess model performance on culturally specific Thai instructions (RQ1), and (2) an instruction-tuning study to measure the effect of native supervision compared to translated datasets (RQ2), including ablation and generalization tests.

These studies require a dataset that supports both evaluation and instruction tuning.
Specifically, we need a culturally grounded test set that enables fine-grained, zero-shot evaluation across task types and domains for RQ1, and a high-quality training set that enables controlled fine-tuning experiments for RQ2.
In response to this need, we introduce WangchanThaiInstruct, a human-authored dataset of Thai instructions covering diverse real-world use cases, explicitly designed to support both types of evaluation. 
We now describe how WangchanThaiInstruct was constructed to support these two studies.

\subsection{WangchanThaiInstruct Dataset Overview}

To support the proposed studies, WangchanThaiInstruct is constructed as a dual-purpose dataset for both zero-shot evaluation and instruction fine-tuning. 
As illustrated in Figure~\ref{fig:figure1}, it comprises 35,014 human-authored instruction-response pairs, split into 28,098 for training and 6,916 for testing.
The dataset is designed to probe model behavior across both general and culturally specific Thai instructions (RQ1) and to provide high-quality supervision for instruction tuning (RQ2).

WangchanThaiInstruct spans four domains—Finance, Legal, Medical, and Retail—each sourced from curated Thai websites to ensure topical relevance and language authenticity. 
It includes seven task types, such as Open QA, Closed QA, Summarization, and Creative writing, selected to reflect a range of real-world applications. 
Each domain covers diverse subtopics (e.g., 18 in Finance, 133 in Legal, 28 in Medical), offering a rich and balanced distribution for both evaluation and generalization analysis.
We provide the full list of subtopics in Figure~\ref{fig:subtopics}.

\begin{figure}[htbp]
\vspace{-3mm}
\centering
\includegraphics[width=0.5\textwidth]{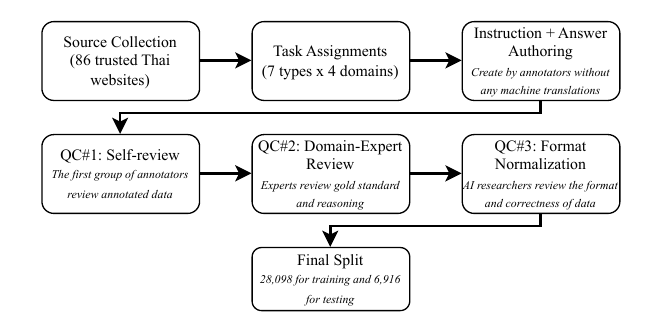}
\vspace{-8mm}
\caption{\label{font-table} The overview of our data collection and annotation pipeline.}
\vspace{-5mm}
\label{fig:pipeline}
\end{figure}

As shown in Figure~\ref{fig:pipeline}, to ensure the dataset’s reliability, we employed a three-stage annotation pipeline with dedicated groups responsible for (1) initial data creation, (2) content verification, and (3) formatting consistency. 
The cultural specificity is annotated by domain experts to enable subgroup analysis in RQ1. 
A detailed account of the collection and annotation process is provided in the following sections.

%
%
%
%

\subsection{Data Collection and Annotation}
\noindent
\textbf{Data Collection}. We collected documents and articles from a diverse set of Thai websites, primarily operated by government agencies and publicly listed companies.
We utilize these documents for annotators to formulate data according to the given document, instead of allowing annotators to come up with a random topic. 
%
In total, we collected 27,351 documents from 86 websites.
During preprocessing, the raw HTML is cleaned by removing unnecessary elements such as HTML tags and images.
We retained only content-rich documents that fell within the top 80\% in terms of token count.
Duplicate or highly similar documents were eliminated using \texttt{mUSE}~\cite{muse}, with a cosine similarity threshold of 0.8.
%
%
A complete list of the source websites is provided in Appendix~\ref{appendix/data_source}.

\noindent
\textbf{Annotation}. Annotators were selected based on two criteria: (i) native fluency in Thai and (ii) successful completion of an annotation exam covering all seven task types.
Only annotators who passed this evaluation were involved in the project.
%
%
For each annotation round, we provided annotators with a sample from a Thai website and a specific task type, where one website could be annotated for multiple tasks across different rounds.
Annotators were instructed to create questions and answers according to the assigned task type.
The annotation guideline and rules are discussed in Appendix~\ref{appendix:first_group_annotators}.

To ensure the quality of the gold standards, annotators were required to provide the reasoning behind each answer (except for creative writing, brainstorming, and summarization tasks), which experts in the field would later verify.
This approach encouraged annotators to ensure they understood the questions and could answer them correctly.
When a website was unsuitable for creating questions or answers (e.g., the provided sample was low quality or the topic was sensitive), annotators were permitted to skip that sample.

\subsection{Quality Control (QC) by Human} \label{subsec:qc_human}
To formulate high-quality data, we implemented three layers of quality control.
In particular, we ask reviewers to check the correctness of the question, gold standard, and formatting for all data in both training and test sets.
We developed an annotation platform, shown in Figure~\ref{fig:annotation_platform.domain_expert} of Appendix~\ref{appendix/annotation_platform.domain_expert}, for annotators to evaluate the instructions.
In addition, we discuss the annotation guidelines and criteria for all annotator groups in Appendix~\ref{appendix:annotator_guideline}.
We describe each aspect as follows:

\noindent
\textbf{Question QC by Annotators}. For the initial quality control, we sampled 10\% of the examples and asked the same annotators to check and edit the questions, instructions, answers, and reasoning for correctness.
%

\noindent
\textbf{Gold Standard QC by Experts}. In this step, we ask domain specialists who work in the target fields (i.e., medical, legal, and finance) to verify the questions and answers from the previous annotation stage.
Each example is reviewed by an expert, and if it is found to be of poor quality in either the question or the answer, or incorrect, those samples are returned to the first step.
When experts reject samples, they are required to provide references and reasons why the samples are incorrect.
We repeated this process until all the data were accepted by experts.
Moreover, we asked experts to add a cultural tag when the sample is Thai-specific content or not, e.g., a question about Thai stocks, Thai medicine, or Thai laws. 
%

\noindent
\textbf{Format QC by AI Researchers}. 
To ensure that WangchanThaiInstruct could serve as high-quality supervision for instruction tuning, we included a dedicated formatting stage led by AI researchers. 
While the first two QC layers focused on content accuracy and domain correctness, this final step emphasized structural consistency with instruction-tuning best practices. 
AI researchers with hands-on experience in training LLMs curated formatting guidelines tailored to each task type (e.g., proper multiple-choice structure, full-sentence closed-form QA, and inclusion of rationale).
A separate group of annotators then implemented these refinements under close supervision. 
This step ensured that the dataset conformed to task-specific conventions expected by modern LLMs, enabling cleaner training and more reliable evaluation.


\noindent
\textbf{Data Leakage}. Since our data has training and test data, and to prevent data leakage, we split the samples by the original document.
Specifically, samples created from the same document, regardless of task type, were categorized in the same data split (either training or test sets).
In addition, for the training data distribution, we have Legal: 64.11\%, Finance: 19.37\%, and Medical: 16.53\%. 
For the test data: Legal: 68.18\%, Finance: 22.07\%, Medical: 9.75\%

%% file: 04_experimental_setup.tex
\section{Experiment Setup}

\subsection{Evaluation Setup} \label{subsec:setup}
Since we provide a train-test split, we evaluate the model under zero-shot settings (the model was never trained on our data) and in-domain settings (the model was trained using our training data).
Moreover, we evaluate those models using out-of-domain benchmarks, such as the Thai LLM leaderboard and the Thai MT-Bench benchmarks, to assess the robustness of models that were trained with our dataset.
We describe each setting as follows.

\begin{table*}[h!]
\large
\centering
\renewcommand{\arraystretch}{1.2}
\resizebox{\textwidth}{!}{%
\begin{tabular}{l|ccccccccccc}
\toprule
 & \multicolumn{1}{c}{\textbf{Brainstorming}} 
 & \multicolumn{2}{c}{\textbf{Classification}}
 & \multicolumn{2}{c}{\textbf{Closed QA}}
 & \multicolumn{1}{c}{\textbf{Creative Writing}}
 & \multicolumn{2}{c}{\textbf{Multiple Choice}}
 & \multicolumn{2}{c}{\textbf{Open QA}}
 & \multicolumn{1}{c}{\textbf{Summarization}} \\
 
\multicolumn{1}{c}{\textbf{Model}}
 & \multicolumn{1}{|c}{\textbf{Fluency}}
 & \multicolumn{1}{c}{\textbf{Accuracy (\%)}}
 & \multicolumn{1}{c}{\textbf{Rating}}
 & \multicolumn{1}{c}{\textbf{Accuracy (\%)}}
 & \multicolumn{1}{c}{\textbf{Rating}}
 & \multicolumn{1}{c}{\textbf{Fluency}}
 & \multicolumn{1}{c}{\textbf{Accuracy (\%)}}
 & \multicolumn{1}{c}{\textbf{Rating}}
 & \multicolumn{1}{c}{\textbf{Accuracy (\%)}}
 & \multicolumn{1}{c}{\textbf{Rating}}
 & \multicolumn{1}{c}{\textbf{Fluency}} \\
 \midrule
\multicolumn{11}{l}{\textbf{Cultural}} \\
Gemini 2.0 & \textbf{7.68} & \textbf{71.50} & \textbf{7.47} & \textbf{96.18} & \textbf{9.20} & \textbf{8.17} & \textbf{62.70} & \textbf{7.02} & \textbf{62.13} & \textbf{6.82} & 8.35\\
Qwen2.5-72B & 7.05 & 63.77 & 6.75 & 93.75 & 9.02 & 7.63 & 54.37 & 6.44 & 45.59 & 5.64 & \textbf{8.60}\\
Llama-3.1-70B & 5.98 & 52.90 & 4.33 & 90.66 & 7.09 & 6.03 & 55.17 & 4.01 & 34.05 & 4.19 & 6.13\\
Sailor2-20B & 6.41 & 62.32 & 6.55 & 92.01 & 9.07 & 6.07 & 58.73 & 6.15 & 44.85 & 5.53 & 7.87\\
Llama-3.1-8B & 4.70 & 34.30 & 4.02 & 79.86 & 6.87 & 5.15 & 40.87 & 4.06 & 15.81 & 3.23 & 6.56\\
Sailor2-8B & 6.50 & 60.87 & 6.38 & 90.97 & 8.91 & 7.00 & 59.92 & 6.59 & 39.34 & 5.24 & 7.99\\
Qwen2.5-7B & 5.95 & 55.07 & 5.84 & 87.85 & 8.32 & 6.71 & 45.63 & 5.59 & 27.94 & 4.24 & 8.22\\
\midrule
\multicolumn{11}{l}{\textbf{General}} \\
Gemini 2.0 & \textbf{7.68} & \textbf{84.38} & \textbf{8.23} & \textbf{97.51} & 9.00 & \textbf{7.95} & \textbf{69.79} & \textbf{7.46} & \textbf{78.53} & \textbf{7.77} & 8.32\\
Qwen2.5-72B & 6.91 & 83.43 & 7.96 & 94.43 & 9.03 & 7.14 & 66.58 & 7.33 & 69.08 & 7.05 & \textbf{8.83}\\
Llama-3.1-70B & 5.92 & 78.43 & 5.86 & 94.40 & 7.38 & 6.07 & 66.02 & 4.71 & 54.65 & 5.38 & 6.14\\
Sailor-2-20B & 6.25 & 83.79 & 7.94 & 95.74 & \textbf{9.13} & 5.78 & 66.71 & 6.79 & 70.14 & 7.12 & 7.93\\
Llama-3.1-8B & 4.90 & 57.99 & 5.38 & 79.70 & 6.92 & 4.83 & 50.13 & 4.82 & 31.24 & 4.10 & 7.05\\
Sailor2-8B & 6.60 & 79.53 & 7.79 & 93.53 & 8.99 & 6.56 & 66.07 & 7.15 & 64.08 & 6.81 & 8.08\\
Qwen2.5-7B & 6.20 & 72.43 & 6.83 & 89.25 & 8.44 & 6.31 & 52.44 & 6.15 & 53.67 & 5.68 & 8.46\\
\bottomrule
\end{tabular}%
}
\vspace{-3mm}
\caption{Comparison of zero-shot performance across cultural and general samples.}
\vspace{-5mm}
\label{tab:main-ood-two-set}
\end{table*}

\noindent \textbf{Test Data Evaluation}: We evaluate existing instruction LLMs without any further training using the same prompt for all models.
We selected well-known instruction LLMs that supported Thai, representing a range of sizes and architectures. The models evaluated include: Gemini 2.0, Qwen2.5 7B and 72B \cite{qwen25}, Llama-3.1 8B and 70B \cite{llama3}, and Sailor2 8B and 20B \cite{sailor}. 
We excluded the GPT family from this evaluation set, as GPT 4.1 was used as the judge model to assess response quality on our benchmark tasks.

\noindent \textbf{Training Data Evaluation}: 
In this setting, we test the robustness of our training data by training base LLMs using our data and testing on our test data and out-of-domain benchmarks.
Given the long-form nature of the outputs in our dataset, we found it more effective to integrate with existing instruction datasets rather than using them as a standalone resource.
Thus, we chose to combine our training dataset with two widely used Thai instruction datasets: \texttt{alpaca-cleaned-52k-th}\footnote{\url{https://huggingface.co/datasets/Thaweewat/alpaca-cleaned-52k-th}} and \texttt{databricks-dolly-15k-th}\footnote{\url{https://huggingface.co/datasets/Thaweewat/databricks-dolly-15k-th}}.
%
%
To ensure fair comparisons, we conduct sample combinations such that the total number of examples remains consistent across setups.
For instance, \texttt{Alpaca 10k} is directly compared with \texttt{Alpaca 5k + ThaInstruct 5k}, maintaining the same total size while varying the composition, where the range is from 2.5k to 30k.
Additionally, we included full-size mixtures—such as \texttt{Alpaca 52k + Thai Instruct 28k} and \texttt{Dolly 15k + Thai Instruct 28k}.
%
%
Our experiments utilized three base models with proven capabilities for Thai and multilingual: {Gemma-2-9B}~\cite{gemma2}, {Llama-3.1-8B}~\cite{llama3}, and {SEA-LIONv2-8B}~\cite{sealion}.
For each model, we maintained consistent hyperparameters, shown in Table~\ref{tab:training hyperparam}, across training runs for fair comparison.

For \emph{out-of-domain benchmarks}, we use out-of-domain datasets, namely the Thai LLM Leaderboard.
Although this benchmark is translation (human verified) and non-cultural data, we want to compare the generalization of models for a fair comparison with previous Thai LLM works.  
We use the default inference setup and codes from the original Thai LLM Leaderboard.
%
%
%

\subsection{Metric}
A key aspect of our benchmark is evaluating the justification and reasoning behind each answer (see reasoning examples in Figure~\ref{fig:reasoning_example}).
Traditional metrics such as BLEU or ROUGE-L fail to capture this dimension, since two correct explanations may differ in wording or format yet still arrive at the same valid conclusion.
%
%
Building on the foundation of MTBench \cite{mtbench}, we utilize LLMs to assess both correctness for tasks with objective answers (e.g., QA tasks) and to assign a 1–10 \textbf{rating} based on the quality of reasoning or justification.
For creative tasks, including brainstorming, creative writing, and summarization, response quality is measured through assessments of \textbf{fluency} and cohesiveness.
For tasks requiring definitive answers, such as question answering and classification, our evaluation is twofold: we verify the factual \textbf{accuracy} and evaluate the accompanying reasoning against human-labeled annotations that justify correct or incorrect responses.
The complete prompts for both categories are shown in Fig.~\ref{fig:prompt}.
Moreover, we study the robustness of existing metrics (i.e., BLEU and ROUGE-L) and our metrics compared to human preferences in Appendix~\ref{subsec:metric_human}.

%% file: 05_results.tex
\section{Experiment Results}

\begin{table*}[h!]
\large
\centering
\renewcommand{\arraystretch}{1.2}
\resizebox{\textwidth}{!}{%
\begin{tabular}{l|ccc|ccc|ccc|ccc}
\toprule
 & \multicolumn{3}{c|}{\textbf{Finance}} 
 & \multicolumn{3}{c|}{\textbf{Medical}}
 & \multicolumn{3}{c|}{\textbf{Retail}}
 & \multicolumn{3}{c}{\textbf{Legal}} \\
\multicolumn{1}{c}{\textbf{Model}}
 & \multicolumn{1}{|c}{\textbf{Fluency}}
 & \multicolumn{1}{c}{\textbf{Accuracy (\%)}}
 & \multicolumn{1}{c|}{\textbf{Rating}}
 & \multicolumn{1}{c}{\textbf{Fluency}}
 & \multicolumn{1}{c}{\textbf{Accuracy (\%)}}
 & \multicolumn{1}{c|}{\textbf{Rating}}
 & \multicolumn{1}{c}{\textbf{Fluency}}
 & \multicolumn{1}{c}{\textbf{Accuracy (\%)}}
 & \multicolumn{1}{c|}{\textbf{Rating}}
 & \multicolumn{1}{c}{\textbf{Fluency}}
 & \multicolumn{1}{c}{\textbf{Accuracy (\%)}}
 & \multicolumn{1}{c}{\textbf{Rating}} \\
\midrule
Gemini 2.0 & \textbf{8.18} & \textbf{79.89} & \textbf{7.82} & 8.19 & \textbf{88.71} & \textbf{8.70} & \textbf{7.88} & \textbf{80.78} & \textbf{7.89} & 8.42 & \textbf{72.35} & \textbf{7.60}\\
Qwen2.5-72B & 7.86 & 76.89 & 7.71 & \textbf{8.69} & 83.84 & 8.18 & 7.53 & 77.10 & 7.77 & \textbf{8.63} & 61.64 & 6.85\\
Llama-3.1-70B & 6.04 & 73.87 & 6.00 & 6.47 & 73.62 & 5.83 & 5.99 & 77.22 & 6.05 & 6.16 & 56.54 & 4.70\\
Sailor2-20B & 6.43 & 76.48 & 7.57 & 8.47 & 85.75 & 8.21 & 6.62 & 77.45 & 7.65 & 8.02 & 62.07 & 6.65\\
Llama-3.1-8B & 5.68 & 57.81 & 5.50 & 7.16 & 51.37 & 5.14 & 5.52 & 59.01 & 5.54 & 6.49 & 40.55 & 4.43\\
Sailor2-8B & 6.96 & 71.15 & 7.41 & 8.55 & 84.07 & 8.19 & 6.99 & 73.64 & 7.54 & 8.17 & 60.20 & 6.64\\
Qwen2.5-7B & 7.12 & 67.26 & 6.80 & 8.41 & 69.15 & 6.90 & 6.93 & 69.58 & 6.94 & 8.25 & 49.73 & 5.77\\
\bottomrule
\end{tabular}%
}
\vspace{-1mm}
\caption{Comparison of zero-shot performance across each domain.}
\vspace{-4mm}
\label{tab:domain-ood}
\end{table*}

\subsection{Test Data Evaluation}
To answer \textbf{RQ1}, we evaluate existing LLMs using our test data without any fine-tuning.
%

\noindent
\textbf{Results.} As shown in Table~\ref{tab:main-ood-two-set}, while Gemini 2.0 outperforms open-source models across several tasks, Qwen2.5-72B demonstrates comparable performance in various tasks. 
For example, in the cultural test set, Qwen2.5-72B outperforms Gemini 2.0 on Summarization, and also outperforms Gemini 2.0 on the Closed QA in the general set.
We also observed that Sailor2-20B and Llama-3.1-70B perform comparably to Gemini 2.0 on several metrics.
These findings suggest promising potential for open-source models to match or exceed the performance of proprietary models like Gemini.

\noindent
\textbf{Discussion.}
From our experimental results, we found that the most challenging tasks are multiple choice and Open QA.
Although previous works demonstrate a high performance for LLMs in Multiple choice~\cite{li-etal-2024-multiple,balepur-etal-2024-artifacts}, the accuracy of this task is low compared to Closed QA or Classification.
This is because our Multiple choice samples require correct reasoning for the answer to be correct. 
This emphasizes the challenge of our datasets; although many previous works demonstrated that Multiple choice datasets were solved by LLMs, our dataset demonstrates a contradiction with the previous works.
%
%

\noindent
\textbf{Reasoning Evaluation}. As shown in Table~\ref{tab:main-ood-two-set}, we evaluate the LLMs’ reasoning against the gold standard using a rating metric.
To quantify the relationship between reasoning quality and correctness, we compute Spearman's rank correlation between rating and accuracy, yielding a strong correlation of 0.78.
This result supports our hypothesis that reasoning quality is closely tied to answer accuracy, where models that provide better reasoning are likely to produce correct answers.

\noindent
\textbf{Cultural vs. General}.
Moreover, we also observe that the performance drops when comparing cultural and general sets.
When we evaluate the accuracy metric, we found that performance dropped in all models and tasks, except for Closed QA.
This emphasizes that the cultural set needs cultural understanding, not only world knowledge like other benchmarks.
In contrast, when evaluating the free-form tasks, i.e., Brainstorming, Creative writing, and Summarization, the performance of these tasks is almost the same for both cultural and general sets.
This is because these tasks do not have an accurate answer; the texts can be answered in a similar form or text.

\subsection{Domain Performance}
Since our dataset spans four distinct domains, LLM performance may vary across them.
To address the dataset gap and challenges (\textbf{RQ1}), we analyze each model’s domain-specific performance using both general and cultural sets from Table~\ref{tab:main-ood-two-set}.

As shown in Table~\ref{tab:domain-ood}, we found that the Legal domain was the most challenging. 
No model achieves over 73\% accuracy in this domain, whereas Gemini surpasses 88\% accuracy in other domains.
This may be due to limited legal data in LLM training, compounded by the fact that laws vary significantly across countries and legal systems, making it harder for LLMs to provide accurate answers.
In contrast, the Medical domain, despite its real-world complexity, yields the highest accuracy, with Gemini achieving 88.71\%.
However, when examining fluency, all domains perform similarly.
This suggests that while generating fluent Thai text is not difficult for these models, producing factually accurate legal responses remains a key challenge.
%
%

\begin{table*}[h!]
\small
\centering
\renewcommand{\arraystretch}{1.2}
\resizebox{\textwidth}{!}{%
\begin{tabular}{l|c|c|cc|ccc}
\toprule
 & \multicolumn{1}{c|}{\textbf{MT Bench}} 
 & \multicolumn{1}{c|}{\textbf{NLU}}
 & \multicolumn{2}{c|}{\textbf{NLG}}
 & \multicolumn{3}{c}{\textbf{WangchanThaiInstruct}} \\
 
\multicolumn{1}{c|}{\textbf{Model}}
 & \multicolumn{1}{c|}{\textbf{Average}}
 & \multicolumn{1}{c|}{\textbf{Accuracy (\%)}}
 & \multicolumn{1}{c}{\textbf{Translation (BLEU)}}
 & \multicolumn{1}{c|}{\textbf{Generation (RougeL)}}
 & \multicolumn{1}{c}{\textbf{Fluency}}
 & \multicolumn{1}{c}{\textbf{Accuracy (\%)}}
 & \multicolumn{1}{c}{\textbf{Rating}} \\
\midrule
\multicolumn{8}{l}{\textbf{Llama-3.1-8B}} \\
Alpaca 52k & 3.04 & 48.48 & \textbf{2.21} & \textbf{9.51} & 3.10 & 26.28 & 3.19\\
Alpaca 52k + WangchanThaiInstruct 28k & \textbf{3.43} & \textbf{50.15} & 2.17 & 8.54 & \textbf{4.52} & \textbf{43.38} & \textbf{4.58}\\
\midrule
Dolly 15k & 2.64 & 42.47 & \textbf{1.60} & 8.10 & 1.85 & 40.34 & 2.63 \\
Dolly 15k + WangchanThaiInstruct 28k & \textbf{2.88} & \textbf{45.56} & 1.28 & \textbf{8.60} & \textbf{4.41} & \textbf{43.51} & \textbf{4.54}\\
\midrule
\multicolumn{8}{l}{\textbf{Gemma-2-9B}} \\
Alpaca 52k & 3.43 & 53.23 & 1.11 & 7.04 & 2.61 & 14.35 & 2.26\\
Alpaca 52k + WangchanThaiInstruct 28k & \textbf{4.61} & \textbf{53.95} & \textbf{1.86} & \textbf{8.05} & \textbf{4.87} & \textbf{55.94} & \textbf{5.20}\\
\midrule
Dolly 15k & \textbf{4.10} & 51.43 & \textbf{1.48} & 7.76 & 1.85 & 40.34 & 2.63\\
Dolly 15k + WangchanThaiInstruct 28k & 3.86 & \textbf{53.88} & 1.47 & \textbf{8.06} & \textbf{4.87} & \textbf{54.44} & \textbf{5.16}\\
\midrule
\multicolumn{8}{l}{\textbf{SEA-LIONv2-8B}} \\
Alpaca 52k & \textbf{4.80} & \textbf{43.94} & 14.59 & \textbf{25.73} & 4.75 & 44.41 & 4.44\\
Alpaca 52k + WangchanThaiInstruct 28k & 4.76 & 43.87 & \textbf{16.40} & 16.51 & \textbf{5.34} & \textbf{51.47} & \textbf{5.33}\\
\midrule
Dolly 15k & 3.57 & \textbf{46.14} & \textbf{14.31} & \textbf{35.37} & 3.24 & 48.13 & 4.15\\
Dolly 15k + WangchanThaiInstruct 28k & \textbf{4.13} & 43.93 & 13.38 & 16.09 & \textbf{5.17} & \textbf{50.52} & \textbf{5.11}\\

\bottomrule
\end{tabular}%
}
\vspace{-1mm}
\caption{Train LLMs using our training data with other training data to improve the robustness of LLMs in out-of-domain and in-domain evaluations. We also compare the \underline{balance data size} in Section~\ref{subsec:balance_data}.}
\vspace{-3mm}
\label{tab:dataset-full-main}
\end{table*}

\subsection{Using Our Training Data}
To answer \textbf{RQ2}, we evaluate the effectiveness of our training data by fine-tuning base LLMs (Llama-3.1, Gemma-2, and SEA-LIONv2) using our data alongside comparison datasets such as Dolly and Alpaca.
%
Additionally, we conduct experiments with both the full dataset and a size-matched subset.
To avoid biased in-domain evaluation, we also assess performance on the Thai LLM Benchmark using MT-Bench, NLU, and NLG datasets.
The full experimental setup is detailed in Section~\ref{subsec:setup}.

\subsubsection{Full Data Comparison}

Table~\ref{tab:dataset-full-main} demonstrates the results of our training data compared with other training data.
We found that, when we use the full dataset for both Dolly/Alpaca and our data, we outperform using only Dolly/Alpaca in 31 out of 42 cases (73\% of cases that we improved the performance).
We observe that we clearly obtain improvements in all cases on in-domain data.
This shows that using our dataset with other data can improve the performance of LLMs, emphasizing the robustness of our training data, which not only improves the in-domain performance but also improves the out-of-domain performance.

Moreover, we also observe consistent improvement for Llama-3.1-8b on the MT Bench and NLU benchmarks.
However, for other base models, we found that there were mixed results in NLG tasks.
This is because NLG tasks are measured using exact match metrics like BLEU and ROUGE-L, which is often ineffective for measuring the performance of LLMs, as we see minor differences in performance of Alpaca/Dolly and our dataset.
%
In contrast, we found that when we use models that are specific to Southeast Asian (SEA) like SEA-LION, the performance difference gap is larger because the model is designed specifically for Thai and other languages in SEA, resulting in a greater improvement when trained in SEA languages.

\subsubsection{Balanced Dataset Study} \label{subsec:balance_data}
To ensure that the observed performance improvement was not merely due to increased dataset size, we matched the sizes of our dataset and Alpaca/Dolly, and conducted experiments using the same benchmark.
%
For example, we compare Alpaca 30k with a mixed dataset of Alpaca 15k and WangchanThaiInstruct 15k, and also experimented with varying total training sizes of 10k, 20k, and 30k to assess the effect of dataset size.
%

As shown in Table~\ref{tab:dataset-ablation-main}, when we split the data equally, we outperform Dolly and Alpaca by 31 out of 42 cases using Llama-3.1-8b, 41 out of 42 cases using Gemma-2-9b, and 28 out of 42 cases using SEA-LIONv2-8B.
Although there were mixed results in NLG datasets, we still outperform competitive datasets in NLU datasets for both Llama-3.1 and Gemma-2.
In addition, the in-domain performance also increases greatly in all cases.
This emphasizes the effectiveness of our training data for in-domain and out-of-domain evaluation regardless of the size of the training data.

%% file: 06_analysis.tex
\subsection{Context Length Analysis}
\begin{figure}[htbp]
\vspace{-1mm}
    \centering
    \includegraphics[width=\columnwidth, trim={0 0 0 0cm}, clip]{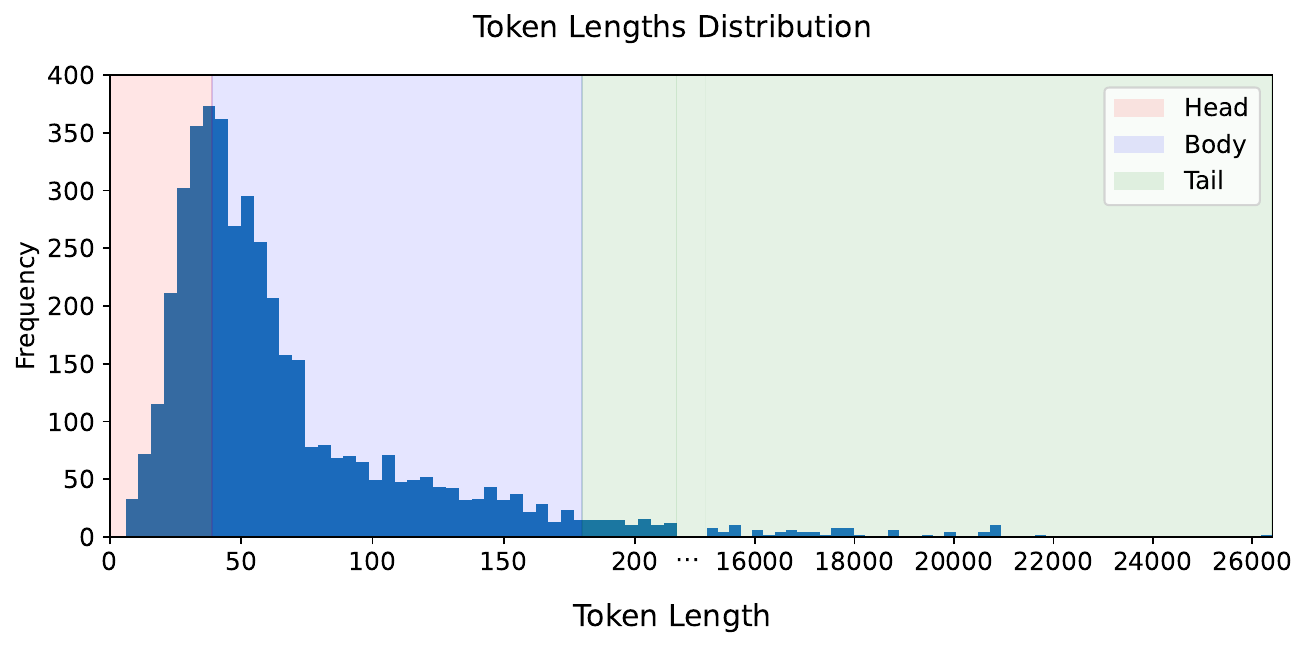}
    \vspace{-6mm}
    \caption{Token length distribution in our dataset.}
    \vspace{-1mm}
    \label{fig:sys-level-assoc}
\end{figure}

Since our test set contains samples with long contexts, ranging from 6 to 26,405 tokens for the test set, we study the gaps and challenges in those samples that contain long contexts.
To facilitate this analysis, we divide the samples into three groups based on token length: head (shortest 20\%), body (middle 40\%), tail (longest 40\%), as illustrated in Figure~\ref{fig:sys-level-assoc}.
We expect robust models to perform consistently regardless of token lengths.

Table~\ref{tab:ana-token-length} presents the performance of each model across the head, body, and tail distributions.
When analyzing the samples with long context, we observe a significant performance drop in the Brainstorming, Multiple choice, Open QA, and Summarization tasks across all models when comparing head and tail groups, with the sole exception of Gemini in Summarization.
Notably, models constrained by shorter context length (i.e., Llama-3.1) or those trained on limited context length data (i.e., Sailor2) exhibit a significant performance drop across both the body and tail distributions.
For example, Sailor2-20B's performance on Summarization dropped from 8.97 to 6.46 points ($\sim$27\% decrease).
%
%
These findings underscore the difficulty of our dataset when handling long inputs, where even state-of-the-art models like Gemini struggle to maintain robust performance.
%





%% file: 07_conclusion.tex
\section{Conclusion}
We introduced WangchanThaiInstruct, a human-authored Thai instruction dataset designed to support evaluation and fine-tuning. 
Our experiments revealed a consistent performance gap in state-of-the-art LLMs when handling Thai context-sensitive inputs, and demonstrated that fine-tuning with WangchanThaiInstruct leads to measurable improvements, supported by ablation studies under size- and format-controlled conditions. 
These findings establish WangchanThaiInstruct as both a benchmark and a development tool for culturally and professionally aligned Thai LLMs.

Beyond the dataset, we contribute a transparent and reproducible process for constructing instruction data that reflects both cultural and domain-specific grounding. 
While WangchanThaiInstruct is language-specific, the underlying methodology is generalizable and can guide similar efforts in other underrepresented languages and application domains.
We make all resources publicly available — including the dataset, evaluation splits, training scripts, and all fine-tuned models, even those that achieve state-of-the-art performance on Thai LLM benchmarks — to establish solid baselines, ensure reproducibility, and support future research focused on culturally and professionally relevant Thai applications.

%% file: appendix/appendix.tex
\section{Measuring Alignment Between Metrics and Human Preferences} \label{subsec:metric_human} 
We evaluated three state-of-the-art LLMs (Gemini 2.5, GPT 4.1, and Claude Sonnet 3.7) to see how closely they aligned with human when ranking model outputs.
To ensure broad coverage, we sampled 1,200 examples uniformly across all domains and task types.
Following previous works~\cite{pavlovic-poesio-2024-effectiveness,movva-etal-2024-annotation}, we had annotators rank the outputs from worst to best, while LLMs produced numerical ratings.
We measured the agreement between human rankings and LLM scores using Kendall $\tau$, which explicitly accounts for ties on either side, which often appeared due to the LLMs producing ratings.
Surprisingly, all three LLMs showed only weak correlation with human preferences as shown in Table~\ref{tab:llm-human-alignment}, indicating that even top-performing models struggle to capture the nuances of Thai cultural and contextual understanding. 
To further validate our judge metric, we computed the human-LLM agreement using precision@k, where we treat the LLM’s top-k choices as predictions and the human top-k as the gold standard. 
We find that the top 1 precision of Gemini is 46.2\% and GPT 4.1 is 48.7\%. In addition, the top 3 precisions of Gemini: 73.4\% and GPT 4.1: 80.3\%. These results confirm that LLM-as-a-judge reliably identifies the best-performing models. 
In the end, we opted for \texttt{GPT 4.1} as the judge since it shows the strongest correlation between LLM and human results.

\begin{table}[htbp]
\centering
\renewcommand{\arraystretch}{1.2}
\scalebox{0.7}{
\begin{tabular}{c|cccc|c}
\toprule
\textbf{Model} & \textbf{Annotator 1} & \textbf{Annotator 2} & \textbf{Annotator 3} & \textbf{Annotator Avg} & \textbf{Judge Time (hr)}\\
\midrule
     Gemini 2.5 & \textbf{0.1322} & \textbf{0.0964} & 0.0156 &  0.1255 & 24\\
     GPT 4.1 & 0.1273 & 0.0933 & 0.0107 & \textbf{0.1259} & 1\\
     Sonnet 3.7 & 0.1044 & 0.0915 & 0.0048 & 0.0962 & 8 \\
     BLEU & -0.0010 & 0.0427 & \textbf{0.0198} & 0.0319 & 0.1\\
     ROUGE-L & -0.0115 & 0.0166 & 0.0083 & 0.0053 & 0.5 \\ \hline
\end{tabular}}
\vspace{-3mm}
\caption{Kendall $\tau$ between annotators and LLM and the judge time per model}
\vspace{-5mm}
\label{tab:llm-human-alignment}
\end{table}

\section{Annotator Guideline} \label{appendix:annotator_guideline} 

\subsection{First Group of Annotators}\label{appendix:first_group_annotators}
We have provided information in each domain as a JSON file in the following structure: Title, Texts (Article), URL (Source link). 
Then, we let the annotator write the instruction and gold standard, where each article will have an instruction task type, such as:
Open QA, Closed QA, Multiple choice, Summarization, Brainstorming, Classification, and Creative Writing.
We also randomly assign a task type for users to perform, according to the following rules:
\begin{itemize}
    \item Medical domain has the most data: randomly assign 1 task type per article.
    \item Finance domain: 2 task types per article.
    \item Retail domain: 2 task types per article.
    \item Legal domain: 2 task types per article.
\end{itemize}
We also control each domain's distribution to have the most balanced distribution of task types. 
We also asked annotators to check the length of each article first. If the content is not long enough or too short, assign only 1 task type to that article, except for the Medical domain, which is already 1 task type per article.

\subsection{Domain Experts}\label{appendix:evaluation_cri}
Criteria are grouped into three categories:

\textbf{C1: Format and Scope.} The output must strictly follow the required structure: “Answer + Elaboration.” The inclusion of “Comparison” and “Conclusion” components is optional. Responses must adhere closely to the given instructions and input, avoiding irrelevant content or exceeding the specified scope. For example, in summarization tasks, the response should accurately reflect all key points from the input without introducing information not present in the original content.

\textbf{C2: Factual Accuracy and Completeness.} The output must be factually accurate, logically sound, and fully address all parts of the instruction.  For example, if asked to recommend practice (e.g., the duration to avoid heavy meals before exercise), but the output fails to provide the underlying rationale for this recommendation, it is considered incomplete and consequently does not meet C2.

\textbf{C3: Instruction/Question Relevance.} This criterion evaluates whether the instruction or question is appropriate, specific, and well-aligned with the expected task type and input. It also considers the question’s relevance to the domain and its suitability for eliciting a meaningful response.
Additionally, experts added relevant tags to each example, with no limit on the number of tags as long as they related to the content.
This process ensured that the gold standards in our dataset were 100\% accurate.

\subsection{AI Researcher Annotators}\label{appendix:third_group_annotators}
As we discussed in Section~\ref{subsec:qc_human}, for this group, we only let them check the format of the instruction and the gold standard.
If they found the error or incorrect format, the sample will be sent back to the first group of annotators, similar to the domain experts' pipeline.

\section{Inference Setup}
All models were evaluated using a temperature of 0.2 and a \texttt{max new tokens} of 1028, with inference performed via the vLLM \footnote{\url{https://github.com/vllm-project/vllm}} engine, with the only exception being Gemini, which was accessed through API calls.

\section{Data Sources}
\label{appendix/data_source}
The data were collected in the form of web page documents from the following websites:
\begin{itemize}
    \raggedright

    \item \textbf{Finance}: bam.co.th, finrwealthbuilder.com, kasikornbank.com, longtunman.com, phillip.co.th, setinvestnow.com, finnomena.com, thestandard.co, brandage.com, brandbuffet.in.th, brandinside.asia, ceochannels.com, marketingoops.com, wealthsolution.co.th.
    \item \textbf{Legal}: cpao.go.th, ddproperty.com, dharmniti.co.th, dplawandservice.com, elt-corp.com, justicechannel.org, khemmapat.org, kobkiat.com, lawsiam.com, lawyerthailand.biz, mkclegal.com, moj.go.th, pdpathailand.com, promsaklawyer.com, saranlaw.com, slawconsult.com
    \item \textbf{Medical}: ambu.or.th, bangkokhospital-chiangrai.com, bumrungrad.com, he02.tci-thaijo.org, w1.med.cmu.ac.th, chulalongkornhospital.go.th, dmh.go.th, dst.or.th, thaiepilepsysociety.com, rama.mahidol.ac.th, gastrothai.net, goodbyeitch.com, haamor.com, idthai.org, manarom.com, medparkhospital.com
    \item \textbf{Retail}: ceochannels.com, marketingoops.com, pnstoretailer.com, techsauce.co, storehub.com, brandage.com, brandbuffet.in.th, brandinside.asia, marketthink.co, readthecloud.co.

\end{itemize}


\section{Training Setup}
All models were trained with LlamaFactory~\cite{llamafactory} with the hyperparameters specified in Table~\ref{tab:training hyperparam}.
\begin{table}[h]
    \centering
    \begin{tabular}{lc}
    \textbf{Hyperparameter}\\
    \toprule
         Learning Rate & $2 \times 10^{-4}$\\
         Learning Rate Schedule & Cosine \\
         Batch Size (effective) & 128 \\
         Max Token Length & 2048 \\
         Warm up Ratio & 0.1 \\
         Epochs & 3 \\ \hline
    \end{tabular}
    \caption{Training Hyperparameters.}
    \label{tab:training hyperparam}
\end{table}

\section{Annotation Platform}
\label{appendix/annotation_platform.domain_expert}
\begin{figure}[!h]
    \centering
    \includegraphics[width=0.45\linewidth]{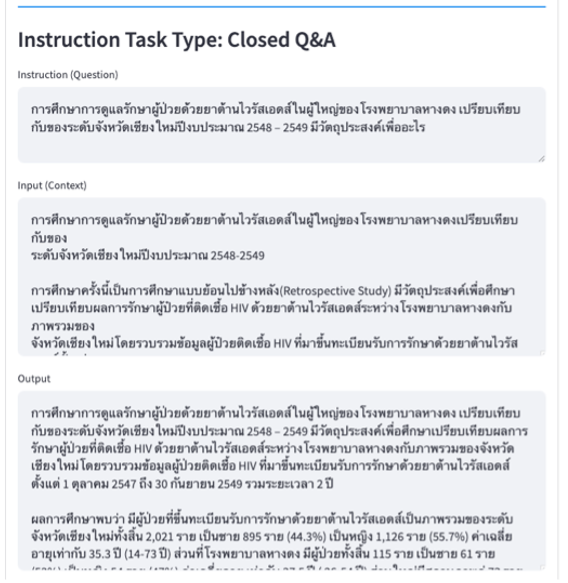}
    \includegraphics[width=0.39\linewidth]
    {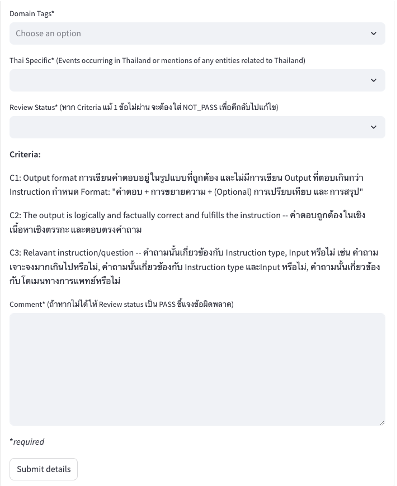}
    \caption{A screenshot of the annotation platform where domain experts can evaluate the quality of instructions based on specific criteria and tag the relevant topics associated with each instruction.}
    \label{fig:annotation_platform.domain_expert}
\end{figure}

As shown in Figure~\ref{fig:annotation_platform.domain_expert}, we developed an annotation platform for the domain experts and QC members to assess the quality of the created instruction stances.

\begin{figure*}[h]
    \small
    \centering
    \begin{mdframed}
        \textbf{Tasks with definitive answers prompt template} \newline
        [Instruction]\newline
        You are to act as an impartial judge tasked with evaluating the quality of a response provided by an AI assistant. 
        The evaluation must focus on two primary criteria: **correctness** and **helpfulness**.\newline\newline
        You will be given:\newline
        - A **user question**.\newline
        - A **reference answer**, which is considered the accurate and authoritative response.\newline
        - An **assistant's answer**, which you must evaluate.\newline\newline
        Your evaluation must begin by determining whether the assistant's answer is **factually correct** and aligns with the information in the reference answer.\newline
        If the assistant's answer contains significant factual inaccuracies, clear contradictions with the reference, or fails to address the core question, mark it as incorrect.\newline
        If the answer is largely accurate and consistent with the reference, mark it as correct.\newline\newline
        Indicate your judgment in the following format:\newline
        Correctness: [[1]]\ (if correct) or Correctness: [[0]]\ (if incorrect)\newline\newline
        Next, provide a **reasoning rating** on a scale from 1 to 10. This rating should reflect how well the assistant's reasoning aligns with the logic, depth, and completeness of the reference answer:\newline
        - A score of 10 means the reasoning is entirely sound, complete, and mirrors the clarity and correctness of the reference.\newline
        - A score of 5 means the reasoning is partially correct or incomplete, but not misleading.\newline
        - A score below 5 indicates flawed or poor reasoning, misunderstanding of the topic, or failure to properly support the answer.\newline
        - A score of 1 means the reasoning is severely flawed or entirely off-topic.\newline\newline
        Use the following format:\newline
        Rating: [[<score>]]\\newline\newline
        [Question]\newline
        {question}\newline\newline
        [The Start of Reference Answer]\newline
        \verb|{ref_answer}|\newline
        [The End of Reference Answer]\newline\newline
        [The Start of Assistant's Answer]\newline
        \verb|{answer}|
    \end{mdframed}
    \begin{mdframed}
        \textbf{Creative tasks prompt template}\newline
        [Instruction]\newline
        Please act as an impartial judge and evaluate the quality of the response provided by an AI assistant to the user question displayed below. Your evaluation should consider clarity, sentence fluency, and cohesion of the writing. You will be given a reference answer and the assistant's answer. Begin your evaluation by comparing the assistant's answer with the reference answer. Identify and correct any mistakes. Be as objective as possible. After providing your explanation, you must rate the response on a scale of 1 to 10 by strictly following this format: [[rating]], for example: Rating: [[5]].\newline\newline
        [Question]\newline
        {question}\newline\newline
        [The Start of Reference Answer]\newline
        \verb|{ref_answer}|\newline
        [The End of Reference Answer]\newline\newline
        [The Start of Assistant's Answer]\newline
        \verb|{answer}|\newline
        [The End of Assistant's Answer]
    \end{mdframed}
    \caption{Adapted MT-Bench prompt templates to fit our benchmark. For tasks with definitive answers we assess both reasoning and correctness of the answers. For creative tasks we assess the clarity, fluency and cohesion of the response.}
    \label{fig:prompt}
\end{figure*}

\begin{figure*}[htbp!]
  \centering
  \includegraphics[width=\textwidth]{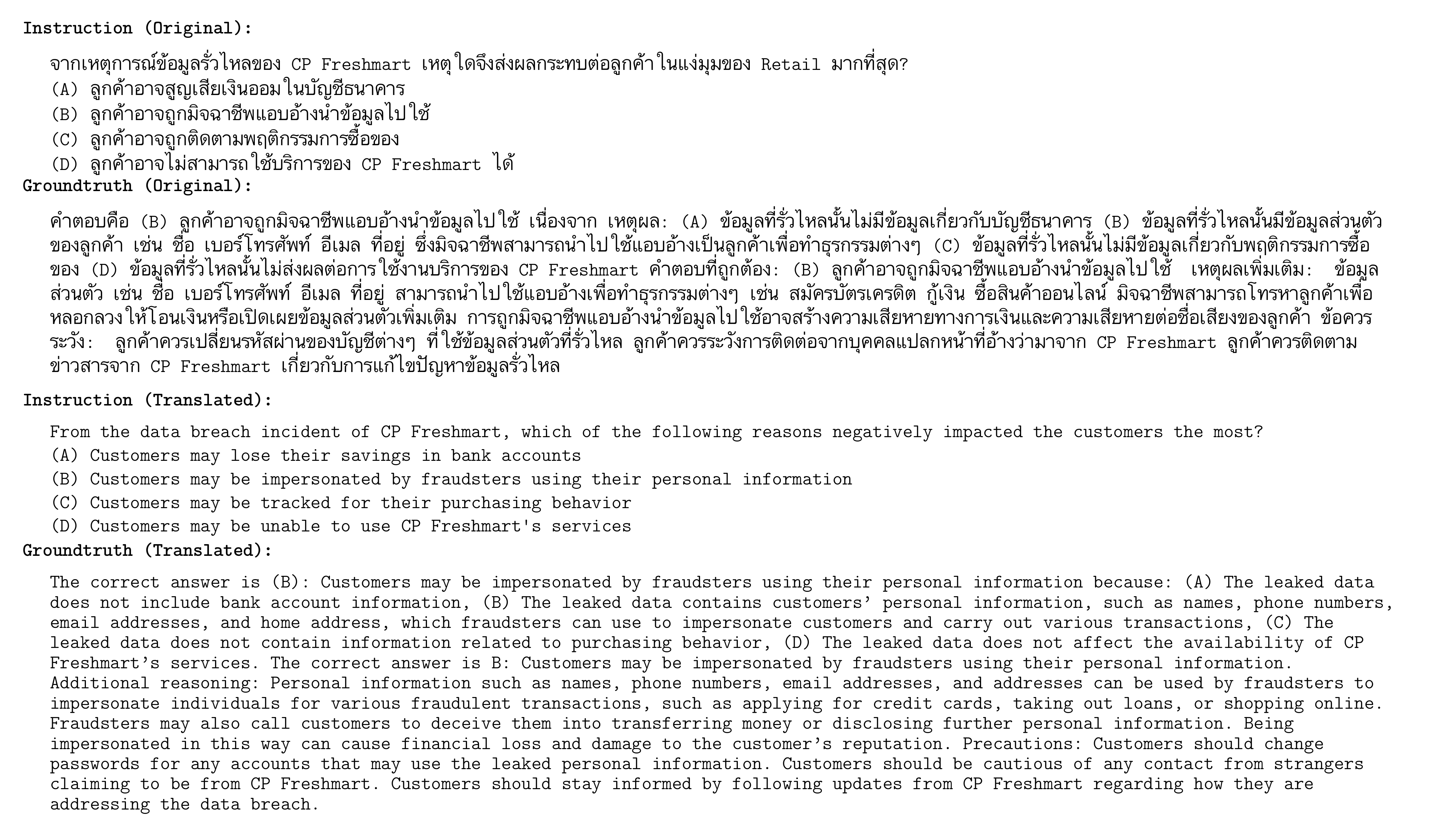}
    \caption{An MCQ example from the dataset showing that each correct answer is followed by the reasoning that arrived towards that answer.}
  \label{fig:reasoning_example}
\end{figure*}

\begin{figure*}
    \small
    \centering
    \begin{mdframed}
        \textbf{Finance Domain Subtopics}:\newline
        Financial Law, Investment Strategies, Personal Financial Management, Asset Management,
        Financial Analysis, Financial Analysis and Financial Economics, Digital Finance, 
        Economic and Financial News, Company Financial Information, Financial Literacy, 
        Financial Market, Financial Market and Financial Products and Services,
        Financial Products and Services, Financial Institutions, Financial Instruments, 
        Financial Technology (FinTech), FinTech and Digital Finance, Financial Economics,
    \end{mdframed}
    \begin{mdframed}
        \textbf{Legal Domain Subtopics}:\newline
        Ministerial Regulation, Maritime Transport and Navigation Law – Merchant Marine and Marine Rescue, Fiscal Law, International Trade Law,
Aviation Law, Gambling Law, Education Law, Education Law – Teachers and Personnel,
Informal Lending Law (Share Lending), Trade Competition Law – Price Control, Loan Law, Loan Law – Usufruct and Consumable Loans,
Immigration Law, Immigration Law – Employment of Foreigners, Family Law, Alcohol Control Law,
Cybersecurity and Privacy Law, Consumer Protection Law, Consumer Protection Law – Product and Service Pricing, Labor Protection Law,
Labor Protection Law – Labor Relations and Compensation, Suretyship Law, Traffic Law, Traffic and Land Transport Law,
Mortgage Law, Pledge Law, Hire of Work Law, Employment Law,
Public Assembly Law, Sales Law, Agency Law, Bill and Cheque Law,
Property Law – Ownership, Property Law – Ownership and Limited Real Rights, Intellectual Property Law, Intellectual Property Law (Copyrights, Patents, Trademarks),
Military Law, Brokerage Law, Juristic Acts and Contracts Law, Juristic Acts and Contracts Law – Unfair Contract Terms,
Juristic Person Law, Current Account and Warehouse Receipt Law, Entertainment Law, Administrative Law,
Administrative Procedure and Administrative Court Establishment Law, Administrative Tort Liability Law, Insurance Law, Social Security Law,
Compromise and Settlement Law, Anti-Money Laundering Law, Deposit Law, Judiciary Act,
Judiciary Act – Royal Decree, Judiciary Act – Jurisdiction Determination between Courts, Energy Law, Merchant Marine Law (Marine Transport and Rescue) – Royal Act,
Merchant Marine Law – Multimodal Transport Royal Act, Tax Law, Inheritance Law, Inheritance Law – Wills,
Narcotics Law, International Law, International Law – Treaties, Constitutional Law,
Transport of Goods and Passengers Law, Tort Law, Law of Evidence, Law of Evidence (Civil and Criminal),
Obligations Law, Obligations Law – Royal Act, Obligations Law – Debt Collection Royal Act, Bankruptcy and Business Reorganization Law,
Building Control, Condominium, Land Allotment and Land Reclamation Law, Building Control Law (draft), Building Control Law (Building Control Royal Act), Building Control Law – Excavation and Land Fill,
Criminal Case Law for Political Office Holders, Computer Crime Law (draft), Computer Crime Law (Computer Act), Tourism Business and Guide Law,
Animal Law, Public Health, Medical and Epidemic Law, Human Rights Law, Environmental Law,
Environmental, Forestry and National Park Law, Health Law, Securities Law, Partnership and Company Law,
Space Law, Real Estate Law, Criminal Law – Petty Offenses, Criminal Law – Public Endangerment (Arson, Flood, etc.),
Criminal Law – Public Endangerment, Criminal Law – Commercial Offenses, Criminal Law – Offenses against the Administration (Officials and Government Positions), Criminal Law – Forgery and Falsification,
Criminal Law – Justice Administration Offenses, Criminal Law – National Security and Terrorism, Criminal Law – Public Peace, Criminal Law – Secret Societies and Criminal Associations,
Criminal Law – Life and Bodily Harm, Criminal Law – Property Offenses, Criminal Law – Theft, Embezzlement, Fraud, Criminal Law – Offenses Related to Corpses,
Criminal Law – Sexual Offenses, Criminal Law – Liberty and Reputation, Weapons, Ammunition, Explosives, Fireworks and Imitation Weapons Law, Hire Purchase Law,
Lease Law, Juvenile and Family Law, Juvenile and Family Law – Royal Act, Juvenile and Family Law – Domestic Violence Protection Royal Act,
Election Law, Labor Law, Exchange and Gift Law, Local Administration Regulation Law,
Industrial Factory and Machinery – Mining Law, Cybersecurity and Personal Data Protection Law (PDPA), General News and Statistics, Local Administration Ordinance,
Basic Legal Knowledge, Supreme Court Judgments, Administrative Court Judgments, Land Code,
Criminal Procedure Code, Civil Procedure Code, Criminal Code, Civil and Commercial Code,
Royal Decree, Royal Decree – Supreme Court President Regulations, Emergency Decree, Organic Act,
Legal Profession and Lawyer Ethics – Complaints, Petitions, Legal Etiquette,

    \end{mdframed}
    \begin{mdframed}
        \textbf{Medical Domain Subtopics}:\newline
        Anatomy, Physiology, Alternative Medicine, Pediatrics,
        Nursing, Ophthalmology, Psychiatry, Dentistry,
        Gynecology, Forensic Medicine, Dermatology, Pathology,
        Epidemiology, Radiology, Anesthesiology, Surgery, 
        Veterinary Medicine,  Public Health, Obstetrics,
        Orthopedics, Internal Medicine, Others, Pharmacology, 
        Transfusion Medicine, Emergency Medicine, Rehabilitation Medicine, 
        Nutrition, Otolaryngology / ENT
    \end{mdframed}
    \caption{Full list of subtopics for Finance, Legal, and Medical domains}
    \label{fig:subtopics}
\end{figure*}

\begin{table*}[hbtp]
\centering
\resizebox{\textwidth}{!}{%
\begin{tabular}{lccccccccccc}
\toprule
 & \multicolumn{1}{c}{\textbf{Brainstorming}} 
 & \multicolumn{2}{c}{\textbf{Classification}}
 & \multicolumn{2}{c}{\textbf{Closed QA}}
 & \multicolumn{1}{c}{\textbf{Creative Writing}}
 & \multicolumn{2}{c}{\textbf{Multiple Choice}}
 & \multicolumn{2}{c}{\textbf{Open QA}}
 & \multicolumn{1}{c}{\textbf{Summarization}} \\
 
\multicolumn{1}{c}{\textbf{Model}}
 & \textbf{Fluency}
 & \textbf{Accuracy (\%)}
 & \textbf{Rating}
 & \textbf{Accuracy (\%)}
 & \textbf{Rating}
 & \textbf{Fluency}
 & \textbf{Accuracy (\%)}
 & \textbf{Rating}
 & \textbf{Accuracy (\%)}
 & \textbf{Rating}
 & \textbf{Fluency} \\
\midrule
\multicolumn{12}{l}{\textbf{Gemini 2.0}} \\
Head & 7.45 & 80.37 & 8.10 & 95.04 & 8.81 & 6.61 & 70.95 & 7.29 & \textbf{77.69} & \textbf{7.69} & \textbf{8.53} \\
Body & 7.61 & \textbf{84.16} & \textbf{8.27} & 97.67 & 9.09 & 8.00 & \textbf{71.78} & \textbf{7.57} & 73.28 & 7.47 & 8.22 \\
Tail & \textbf{7.86} & 80.24 & 7.89 & \textbf{97.87} & \textbf{9.13} & \textbf{8.69} & 62.84 & 7.16 & 75.00 & 7.58 & 8.32 \\
\midrule
\multicolumn{12}{l}{\textbf{Qwen2.5-72B}} \\
Head & 6.79 & 78.08 & 7.63 & 92.37 & 8.98 & 5.93 & 66.67 & 7.26 & 63.75 & 6.64 & \textbf{9.27} \\
Body & 6.90 & \textbf{80.85} & \textbf{7.82} & \textbf{94.94} & 9.00 & 7.30 & \textbf{67.88} & \textbf{7.36} & \textbf{65.79} & \textbf{6.78} & 8.91 \\
Tail & \textbf{7.00} & 79.02 & 7.67 & 94.58 & \textbf{9.07} & \textbf{7.75} & 57.70 & 6.79 & 61.75 & 6.74 & 8.39 \\
\midrule
\multicolumn{12}{l}{\textbf{Llama-3.1-70B}} \\
Head & 5.97 & 70.48 & 5.61 & 88.37 & 6.82 & 5.04 & \textbf{67.44} & \textbf{5.07} & 39.50 & 4.53 & \textbf{6.89} \\
Body & \textbf{6.01} & \textbf{72.60} & \textbf{5.86} & 93.42 & 7.27 & 6.04 & 64.89 & 4.47 & 47.58 & 4.91 & 6.14 \\
Tail & 5.82 & 71.69 & 5.01 & \textbf{95.79} & \textbf{7.55} & \textbf{6.51} & 57.95 & 4.20 & \textbf{53.63} & \textbf{5.37} & 5.80 \\
\midrule
\multicolumn{12}{l}{\textbf{Sailor2-20B}} \\
Head & 6.31 & 78.54 & 7.67 & 93.89 & 9.08 & 5.53 & 64.29 & 6.29 & 63.35 & 6.71 & \textbf{8.97} \\
Body & 6.13 & 78.49 & \textbf{7.68} & \textbf{96.89} & \textbf{9.25} & \textbf{5.89} & \textbf{65.21} & 6.65 & 63.16 & 6.70 & 8.84 \\
Tail & \textbf{6.36} & \textbf{81.22} & 7.64 & 89.94 & 8.53 & 5.86 & 64.55 & \textbf{6.80} & \textbf{66.45} & \textbf{6.86} & 6.46 \\
\midrule
\multicolumn{12}{l}{\textbf{Llama-3.1-8B}} \\
Head & 4.70 & 49.32 & 4.96 & 77.48 & 6.76 & 3.72 & 45.71 & 4.17 & 20.72 & 3.34 & \textbf{7.21} \\
Body & 4.85 & 53.43 & \textbf{5.33} & \textbf{83.07} & \textbf{7.13} & 4.89 & \textbf{49.39} & 4.63 & 24.70 & 3.73 & 7.07 \\
Tail & \textbf{5.02} & \textbf{55.37} & 4.97 & 77.56 & 6.76 & \textbf{5.44} & 47.43 & \textbf{4.88} & \textbf{34.83} & \textbf{4.39} & 6.68 \\
\midrule
\multicolumn{12}{l}{\textbf{Sailor2-8B}} \\
Head & 6.50 & 70.78 & 7.22 & 91.98 & 8.96 & 5.88 & 63.81 & 6.81 & \textbf{64.14} & \textbf{6.60} & \textbf{9.20} \\
Body & 6.55 & 76.60 & \textbf{7.62} & \textbf{93.77} & \textbf{9.05} & 6.62 & \textbf{65.94} & \textbf{7.12} & 55.47 & 6.35 & 9.03 \\
Tail & \textbf{6.67} & \textbf{77.80} & 7.55 & 87.04 & 8.34 & \textbf{6.95} & 63.57 & 7.00 & 58.76 & 6.50 & 6.52 \\
\midrule
\multicolumn{12}{l}{\textbf{Qwen2.5-7B}} \\
Head & 5.96 & \textbf{69.41} & 6.62 & 87.40 & 8.33 & 5.09 & 50.00 & 5.90 & 39.44 & 4.75 & \textbf{8.62} \\
Body & 6.00 & 69.03 & \textbf{6.65} & \textbf{89.49} & 8.41 & 6.46 & \textbf{55.47} & \textbf{6.30} & 47.98 & 5.35 & 8.43 \\
Tail & \textbf{6.48} & 68.78 & 6.63 & 89.17 & \textbf{8.46} & \textbf{6.92} & 46.45 & 5.78 & \textbf{52.35} & \textbf{5.71} & 8.29 \\

\bottomrule
\end{tabular}%
}
\caption{Token length analysis with bolded scores representing best scores in each group}
\label{tab:ana-token-length}
\end{table*}

\begin{table*}[h]
\small
\centering
\resizebox{\textwidth}{!}{%
\begin{tabular}{l|c|c|cc|ccc}
\toprule
 & \multicolumn{1}{c|}{\textbf{MT Bench}} 
 & \multicolumn{1}{c|}{\textbf{NLU}}
 & \multicolumn{2}{c|}{\textbf{NLG}}
 & \multicolumn{3}{c}{\textbf{WangchanThaiInstruct}} \\
 
\multicolumn{1}{c|}{\textbf{Model}}
 & \multicolumn{1}{c|}{\textbf{Average}}
 & \multicolumn{1}{c|}{\textbf{Accuracy (\%)}}
 & \multicolumn{1}{c}{\textbf{Translation (BLEU)}}
 & \multicolumn{1}{c|}{\textbf{Generation (RougeL)}}
 & \multicolumn{1}{c}{\textbf{Fluency}}
 & \multicolumn{1}{c}{\textbf{Accuracy (\%)}}
 & \multicolumn{1}{c}{\textbf{Rating}} \\
\midrule
\multicolumn{8}{l}{\textbf{Llama-3.1-8B}} \\
Alpaca 5k + WangchanThaiInstruct 5k & 3.00 & \textbf{47.22} & 3.12 & 8.59 & \textbf{4.08} & \textbf{39.84} & \textbf{4.16} \\
Alpaca 10k & \textbf{3.05} & 46.54 & \textbf{4.08} & \textbf{11.05} & 3.36 & 28.39 & 3.33 \\
\midrule
Alpaca 10k + WangchanThaiInstruct 10k & \textbf{3.07} & 46.47 & 2.43 & 8.54 & \textbf{4.21} & \textbf{42.31} & \textbf{4.39} \\
Alpaca 20k & 2.75 & \textbf{47.31} & \textbf{2.79} & \textbf{9.14} & 2.77 & 22.32 & 2.94 \\
\midrule
Alpaca 15k + WangchanThaiInstruct 15k &\textbf{3.26} & 44.65 & 1.86 & 8.58 & \textbf{4.35} & \textbf{42.16} & \textbf{4.46} \\
Alpaca 30k & 2.88 & \textbf{47.67} & \textbf{3.47} & \textbf{9.65} & 2.83 & 21.83 & 2.95 \\
\midrule
Dolly 2.5k + WangchanThaiInstruct 2.5k & \textbf{2.40} & \textbf{46.43} & \textbf{3.75} & 8.72 & \textbf{3.57} & \textbf{35.93} & \textbf{3.72} \\
Dolly 5k & 1.88 & 42.87 & 1.98 & \textbf{9.55} & 1.75 & 22.70 & 2.19 \\
\midrule
Dolly 5k + WangchanThaiInstruct 5k & \textbf{2.28} & \textbf{46.43} & \textbf{1.36} & 8.55 & \textbf{3.85} & \textbf{37.89} & \textbf{3.98} \\
Dolly 10k & 1.99 & 42.74 & 1.35 & \textbf{8.96} & 1.69 & 22.35 & 2.14 \\
\midrule
Dolly 7.5k + WangchanThaiInstruct 7.5k & 2.31 & \textbf{46.37} & 1.48 & \textbf{8.59} & \textbf{3.96} & \textbf{39.63} & \textbf{4.11} \\
Dolly 15k & \textbf{2.64} & 42.47 & \textbf{1.60} & 8.10 & 1.69 & 22.21 & 2.16 \\
\midrule
\multicolumn{8}{l}{\textbf{Gemma-2-9B}} \\
Alpaca 5k + WangchanThaiInstruct 5k & \textbf{4.25} & \textbf{53.70} & \textbf{2.25} & \textbf{8.14} & \textbf{4.85} & \textbf{54.24} & \textbf{5.17} \\
Alpaca 10k & 3.98 & 51.71 & 1.39 & 6.84 & 4.00 & 46.62 & 4.26 \\
\midrule
Alpaca 10k + WangchanThaiInstruct 10k & 4.02 & \textbf{53.81} & \textbf{2.02} & \textbf{8.09} & \textbf{4.97} & \textbf{55.33} & \textbf{5.30} \\
Alpaca 20k & \textbf{4.14} & 52.40 & 1.45 & 6.95 & 3.53 & 38.07 & 3.90 \\
\midrule
Alpaca 15k + WangchanThaiInstruct 15k & \textbf{4.20} & \textbf{53.49} & \textbf{1.98} & \textbf{8.02} & \textbf{5.14} & \textbf{56.67} & \textbf{5.49} \\
Alpaca 30k & 3.79 & 52.41 & 1.1 & 5.73 & 3.25 & 32.71 & 3.43 \\
\midrule
Dolly 2.5k + WangchanThaiInstruct 2.5k & \textbf{3.66} & \textbf{54.62} & \textbf{1.75} & \textbf{8.07} & \textbf{4.30} & \textbf{51.86} & \textbf{4.84} \\
Dolly 5k & 2.59 & 53.34 & 1.39 & 7.58 & 1.71 & 42.35 & 2.45 \\
\midrule
Dolly 5k + WangchanThaiInstruct 5k & \textbf{3.99} & \textbf{53.50} & \textbf{1.54} & \textbf{8.12} & \textbf{4.59} & \textbf{54.31} & \textbf{5.08} \\
Dolly 10k & 2.70 & 51.98 & 1.52 & 7.58 & 1.81 & 43.68 & 2.74 \\
\midrule
Dolly 7.5k + WangchanThaiInstruct 7.5k & \textbf{4.13} & \textbf{53.34} & \textbf{1.63} & \textbf{8.12} & \textbf{4.72} & \textbf{55.09} & \textbf{5.24} \\
Dolly 15k & 4.10 & 51.43 & 1.48 & 7.76 & 1.85 & 40.34 & 2.63 \\
\midrule
\multicolumn{8}{l}{\textbf{SEA-LIONv2-8B}} \\
Alpaca 5k + WangchanThaiInstruct 5k & 4.52 & \textbf{43.76} & \textbf{34.47} & 19.39 & \textbf{5.62} & \textbf{52.84} & \textbf{5.57} \\
Alpaca 10k & \textbf{4.54} & 43.31 & 28.01 & \textbf{25.35} & 4.61 & 48.88 & 4.73 \\
\midrule
Alpaca 10k + WangchanThaiInstruct 10k & 4.55 & \textbf{44.66} & 24.00 & 17.55 & \textbf{5.72} & \textbf{53.93} & \textbf{5.70} \\
Alpaca 20k & \textbf{4.74} & 43.98 & \textbf{24.22} & \textbf{25.82} & 4.73 & 49.32 & 4.53 \\
\midrule
Alpaca 15k + WangchanThaiInstruct 15k & 4.44 & \textbf{44.51} & \textbf{20.58} & 16.31 & \textbf{5.54} & \textbf{53.94} & \textbf{5.61} \\
Alpaca 30k & \textbf{4.60} & 42.96 & 15.58 & \textbf{25.68} & 5.11 & 49.66 & 4.78 \\
\midrule
Dolly 2.5k + WangchanThaiInstruct 2.5k & \textbf{4.25} & 44.89 & \textbf{36.60} & 26.82 & \textbf{5.10} & \textbf{50.25} & \textbf{5.28} \\
Dolly 5k & 3.69 & \textbf{45.88} & 25.53 & \textbf{35.66} & 3.46 & 48.04 & 4.11 \\
\midrule
Dolly 5k + WangchanThaiInstruct 5k & \textbf{4.21} & 44.30 & \textbf{15.64} & 23.72 & \textbf{5.31} & \textbf{51.25} & \textbf{5.42} \\
Dolly 10k & 3.83 & \textbf{46.57} & 14.40 & \textbf{37.35} & 3.09 & 48.61 & 4.04 \\
\midrule
Dolly 7.5k + WangchanThaiInstruct 7.5k & \textbf{4.31} & 45.31 & 13.54 & 22.00 & \textbf{5.54} & \textbf{53.81} & \textbf{5.57} \\
Dolly 15k & 3.57 & \textbf{46.14} & \textbf{14.31} & \textbf{35.37} & 3.24 & 48.13 & 4.15 \\
\bottomrule
\end{tabular}%
}
\caption{Dataset ablation: in-domain distribution with data size ablations.}
\label{tab:dataset-ablation-main}
\end{table*}

\input{appendix/tables}

%% file: appendix/tables.tex
\begin{table*}[hbtp]
\centering
\resizebox{\textwidth}{!}{%
\begin{tabular}{l|cccccccc|c}
\toprule
\textbf{Model} & \textbf{Social Science} & \textbf{Extraction} & \textbf{Math} & \textbf{Reasoning} & \textbf{STEM} & \textbf{Writing} & \textbf{Coding} & \textbf{Roleplay} & \textbf{Average} \\
\midrule
\multicolumn{10}{l}{\textbf{Llama 3.1 8B}} \\
Alpaca 5k + ThaiInstruct 5k & 2.9 & 2.85 & 2.2 & 2.3 & 3.7 & 3 & 3.68 & 3.35 & 3.00 \\
Alpaca 10k & 3.65 & 2.35 & 2 & 1.9 & 3.25 & 3.45 & 3.86 & 3.9 & 3.05 \\
\midrule
Alpaca 10k + ThaiInstruct 10k & 4.25 & 2.15 & 2.15 & 2.45 & 3.5 & 3.35 & 3.14 & 3.55 & 3.07 \\
Alpaca 20k & 3.4 & 1.8 & 1.35 & 1.85 & 3.4 & 2.95 & 3.23 & 4.05 & 2.75 \\
\midrule
Alpaca 15k + ThaiInstruct 15k & 3.65 & 3 & 2.05 & 2.9 & 3.5 & 3 & 3.95 & 4 & 3.26 \\
Alpaca 30k & 3.3 & 1.8 & 1.9 & 1.95 & 3 & 3.15 & 3.77 & 4.15 & 2.88 \\
\midrule
Dolly 2.5k + ThaiInstruct 2.5k & 2.65 & 3 & 1.4 & 1.8 & 2.6 & 2.75 & 2.5 & 2.5 & 2.40 \\
Dolly 5k & 2.45 & 1.85 & 1.3 & 1.8 & 2 & 1.65 & 2 & 1.95 & 1.88 \\
\midrule
Dolly 5k + ThaiInstruct 5k & 3.25 & 2.95 & 1.6 & 1.85 & 1.7 & 2.7 & 2.36 & 1.8 & 2.28 \\
Dolly 10k & 2.15 & 2.05 & 1.25 & 1.7 & 1.85 & 2.35 & 3 & 1.6 & 1.99\\
\midrule
Dolly 7.5k + ThaiInstruct 7.5k & 3.9 & 2.5 & 1.1 & 2.1 & 2.45 & 2.5 & 1.86 & 2.1 & 2.31 \\
Dolly 15k & 3.35 & 3.4 & 1.5 & 2.9 & 2.5 & 1.7 & 3 & 2.8 & 2.64 \\
\midrule
\multicolumn{10}{l}{\textbf{Gemma 2 9B}} \\
Alpaca 5k + ThaiInstruct 5k & 6 & 3.9 & 3.05 & 3.75 & 4.1 & 4.05 & 3.64 & 5.5 & 4.25 \\
Alpaca 10k & 5.4 & 2.75 & 2.95 & 2.85 & 5.05 & 4.2 & 3.32 & 5.35 & 3.98 \\
\midrule
Alpaca 10k + ThaiInstruct 10k & 6 & 2.95 & 3.3 & 4.3 & 4.1 & 3.6 & 3.18 & 4.75 & 4.02 \\
Alpaca 20k & 6 & 2.3 & 4.15 & 3.7 & 4.65 & 4.3 & 3.59 & 4.4 & 4.14 \\
\midrule
Alpaca 15k + ThaiInstruct 15k & 6.05 & 2.8 & 3.3 & 3.95 & 4.3 & 4 & 4.18 & 5 & 4.20 \\
Alpaca 30k & 4.65 & 2.45 & 3.85 & 2.55 & 4.75 & 3.25 & 3.77 & 5.05 & 3.79 \\
\midrule
Dolly 2.5k + ThaiInstruct 2.5k & 5.2 & 2.75 & 3.15 & 3.75 & 3.65 & 3.95 & 2.85 & 4 & 3.66 \\
Dolly 5k & 3.2 & 2.4 & 1.6 & 2.8 & 3.4 & 2.15 & 2.05 & 3.15 & 2.59 \\
\midrule
Dolly 5k + ThaiInstruct 5k & 5.85 & 3 & 3 & 4.2 & 4.2 & 4 & 3.32 & 4.35 & 3.99 \\
Dolly 10k & 3.4 & 1.75 & 1.65 & 3.05 & 3.35 & 3.1 & 2.23 & 3.1 & 2.70 \\
\midrule
Dolly 7.5k + ThaiInstruct 7.5k & 6.3 & 3.9 & 3.2 & 3.95 & 4.2 & 4 & 3.18 & 4.3 & 4.13 \\
Dolly 15k & 5.85 & 4.55 & 3.5 & 3.75 & 4.55 & 3.6 & 2.77 & 4.25 & 4.10 \\
\midrule
\multicolumn{10}{l}{\textbf{SEA-LIONv2-8B}} \\
Alpaca 5k + ThaiInstruct 5k & 6.05 & 3.95 & 3.3 & 3.65 & 5.8 & 5.15 & 3.68 & 4.6 & 4.52 \\
Alpaca 10k & 6.25 & 3.9 & 2.4 & 3.35 & 4.95 & 5.55 & 4.27 & 5.65 & 4.54 \\
\midrule
Alpaca 10k + ThaiInstruct 10k & 7.25 & 3.25 & 3.35 & 3.95 & 5.2 & 4.65 & 3.18 & 5.6 & 4.55 \\
Alpaca 20k & 6.8 & 3.35 & 3.4 & 4.45 & 4.95 & 5.2 & 4.18 & 5.55 & 4.74 \\
\midrule
Alpaca 15k + ThaiInstruct 15k & 5.65 & 3.65 & 3 & 3.75 & 5.45 & 4.75 & 3.59 & 5.55 & 4.44 \\
Alpaca 30k & 7 & 3.5 & 3.45 & 4.1 & 5.5 & 4.7 & 3.36 & 5.15 & 4.60 \\
\midrule
Dolly 2.5k + ThaiInstruct 2.5k & 5.35 & 4.25 & 3.1 & 3.5 & 4.75 & 4.75 & 3.68 & 4.6 & 4.25 \\
Dolly 5k & 5.1 & 4.15 & 1.75 & 3.8 & 4.45 & 3.3 & 3.14 & 3.8 & 3.69 \\
\midrule
Dolly 5k + ThaiInstruct 5k & 5.9 & 4.3 & 2.25 & 4.05 & 4.45 & 4.65 & 3.41 & 4.65 & 4.21 \\
Dolly 10k & 5.15 & 2.75 & 2.5 & 3.9 & 4.95 & 3.6 & 3.18 & 4.6 & 3.83 \\
\midrule
Dolly 7.5k + ThaiInstruct 7.5k & 5.75 & 4.45 & 2.2 & 3.45 & 5 & 4.95 & 3.64 & 5.05 & 4.31 \\
Dolly 15k & 5 & 3.55 & 2.1 & 3.7 & 4.2 & 3.2 & 3.18 & 3.6 & 3.57 \\
\bottomrule
\end{tabular}}
\label{tab:dataset-ablation-mtbench}
\caption{Dataset ablation: Thai MT Bench}
\end{table*}

\begin{table*}[hbtp]
\centering
\resizebox{\textwidth}{!}{%
\begin{tabular}{l|cccccccc|c}
\toprule
\textbf{Model} & \textbf{Social Science} & \textbf{Extraction} & \textbf{Math} & \textbf{Reasoning} & \textbf{STEM} & \textbf{Writing} & \textbf{Coding} & \textbf{Roleplay} & \textbf{Average} \\
\midrule
\multicolumn{10}{l}{\textbf{Llama-3.1-8B}} \\
Alpaca 52k + ThaiInstruct 28k & 4.1 & 2.15 & 1.85 & 2.8 & 3.55 & 2.95 & 3.82 & 4.15 & 3.43 \\
Alpaca 52k & 3.4 & 2.05 & 1.65 & 1.95 & 4.1 & 3.45 & 3.55 & 4.2 & 3.04 \\
\midrule
Dolly 15k + ThaiInstruct 28k & 3.75 & 3.5 & 1.65 & 2.3 & 2.75 & 2.55 & 2.64 & 2.85 & 2.88 \\
Dolly 15k & 3.35 & 3.4 & 1.5 & 2.9 & 2.5 & 1.7 & 3 & 2.8 & 2.64 \\
\midrule
\multicolumn{10}{l}{\textbf{Gemma-2-9B}} \\
Alpaca 52k + ThaiInstruct 28k & 6.6 & 4.0 & 3.5 & 3.4 & 4.5 & 4.9 & 4.10 & 4.95 & 4.61 \\
Alapaca 52k & 3.6 & 2.35 & 3.8 & 2.3 & 4.7 & 3.4 & 3.5 & 3.8 & 3.43\\
\midrule
Dolly 15k + ThaiInstruct 28k & 5.8 & 3.25 & 3.4 & 3.75 & 4.75 & 3.75 & 2.95 & 3.25 & 3.86 \\
Dolly 15k & 5.85 & 4.55 & 3.5 & 3.75 & 4.55 & 3.6 & 2.77 & 4.25 & 4.10 \\
\midrule
\multicolumn{10}{l}{\textbf{SEA-LIONv2-8B}} \\
Alpaca 52k + ThaiInstruct 28k & 6.35 & 4.05 & 2.7 & 3.7 & 6.2 & 5.25 & 4.64 & 5.15 & 4.76 \\
Alpaca 52k & 6.75 & 3.65 & 3.25 & 4.3 & 5.7 & 4.7 & 4.41 & 5.6 & 4.80\\
\midrule
Dolly 15k + ThaiInstruct 28k & 6.15 & 3.7 & 2.2 & 4.0 & 4.85 & 4.25 & 4.14 & 3.75 & 4.13 \\
Dolly 15k & 5 & 3.55 & 2.1 & 3.7 & 4.2 & 3.2 & 3.18 & 3.6 & 3.57 \\
\bottomrule
\end{tabular}}
\caption{Full dataset: Thai MT Bench results}
\label{tab:full-dataset-mtbench}
\end{table*}

\begin{table*}[ht]
\centering
\resizebox{\textwidth}{!}{%
\begin{tabular}{l|cccccc|c}
\toprule
\textbf{Model} & \textbf{Belebele} & \textbf{M3 Exam} & \textbf{Thai Exam} & \textbf{Wisesight} & \textbf{xCopa} & \textbf{xNLI} & \textbf{Average} \\
\midrule
\multicolumn{8}{l}{\textbf{Llama-3.1-8B}} \\
Alpaca 5k + ThaiInstruct 5k & 65.89 & 39.02 & 36.64 & 39.12 & 69.60 & 33.07 & 47.22 \\
Alpaca 10k & 63.33 & 40.54 & 33.27 & 45.53 & 66.60 & 29.98 & 46.54 \\
\midrule
Alpaca 10k + ThaiInstruct 10k & 63.78 & 38.70 & 35.04 & 37.55 & 70.40 & 33.37 & 46.47 \\
Alpaca 20k & 63.11 & 40.36 & 33.45 & 50.51 & 66.20 & 30.26 & 47.31 \\
\midrule
Alpaca 15k + ThaiInstruct 15k & 61.56 & 38.28 & 33.10 & 35.75 & 66.20 & 33.01 & 44.65 \\
Alpaca 30k & 62.33 & 39.39 & 32.39 & 50.17 & 69.20 & 32.51 & 47.67 \\
\midrule
Dolly 2.5k + ThaiInstruct 2.5k & 65.11 & 40.68 & 35.58 & 36.20 & 68.00 & 33.03 & 46.43 \\
Dolly 5k & 59.44 & 35.01 & 29.73 & 37.18 & 63.00 & 32.85 & 42.87 \\
\midrule
Dolly 5k + ThaiInstruct 5k & 65.11 & 40.68 & 35.58 & 36.20 & 68.00 & 33.03 & 46.43 \\
Dolly 10k & 60.78 & 35.33 & 28.85 & 35.19 & 63.80 & 32.51 & 42.74 \\
\midrule
Dolly 7.5k + ThaiInstruct 7.5k & 65.11 & 39.44 & 36.99 & 35.04 & 68.60 & 33.03 & 46.37 \\
Dolly 15k & 60.67 & 35.93 & 31.50 & 34.71 & 60.00 & 32.00 & 42.47 \\
\midrule
\multicolumn{8}{l}{\textbf{Gemma-2-9B}} \\
Alpaca 5k + ThaiInstruct 5k & 77.11 & 50.05 & 46.55 & 37.36 & 77.60 & 33.53 & 53.70 \\
Alpaca 10k & 76.89 & 48.11 & 46.73 & 33.17 & 71.00 & 34.37 & 51.71 \\
\midrule
Alpaca 10k + ThaiInstruct 10k & 77.56 & 49.86 & 44.78 & 36.80 & 80.20 & 33.65 & 53.81 \\
Alpaca 20k & 78.56 & 47.92 & 47.43 & 37.18 & 68.00 & 35.33 & 52.40 \\
\midrule
Alpaca 15k + ThaiInstruct 15k & 78.56 & 49.82 & 46.73 & 34.07 & 77.20 & 34.59 & 53.49 \\
Alpaca 30k & 77.44 & 48.25 & 46.73 & 40.51 & 66.80 & 34.71 & 52.41 \\
\midrule
Dolly 2.5k + ThaiInstruct 2.5k & 80.22 & 49.58 & 46.90 & 33.88 & 82.60 & 34.53 & 54.62 \\
Dolly 5k & 78.33 & 49.08 & 46.37 & 33.28 & 78.60 & 34.37 & 53.34 \\
\midrule
Dolly 5k + ThaiInstruct 5k & 79.22 & 49.63 & 46.37 & 33.02 & 79.40 & 33.37 & 53.50 \\
Dolly 10k & 78.00 & 48.66 & 43.89 & 32.31 & 75.60 & 33.41 & 51.98 \\
\midrule
Dolly 7.5k + ThaiInstruct 7.5k & 80.44 & 50.14 & 46.73 & 31.30 & 81.60 & 33.37 & 53.34 \\
Dolly 15k & 75.11 & 47.32 & 43.36 & 33.10 & 71.60 & 38.10 & 51.43 \\
\midrule
\multicolumn{8}{l}{\textbf{SEA-LIONv2-8B}} \\
Alpaca 5k + ThaiInstruct 5k & 57.56 & 40.45 & 36.81 & 32.98 & 61.40 & 33.35 & 43.76 \\
Alpaca 10k & 56.11 & 42.02 & 38.23 & 29.88 & 59.60 & 34.01 & 43.31 \\
\midrule
Alpaca 10k + ThaiInstruct 10k & 57.22 & 39.02 & 34.69 & 33.43 & 70.20 & 33.37 & 44.66 \\
Alpaca 20k & 58.33 & 42.62 & 38.76 & 27.59 & 62.60 & 33.99 & 43.98 \\
\midrule
Alpaca 15k + ThaiInstruct 15k & 57.44 & 39.81 & 36.64 & 35.38 & 64.40 & 33.39 & 44.51 \\
Alpaca 30k & 57.44 & 42.39 & 36.99 & 27.07 & 59.60 & 34.25 & 42.96 \\
\midrule
Dolly 2.5k + ThaiInstruct 2.5k & 57.11 & 42.62 & 38.76 & 36.05 & 61.40 & 33.41 & 44.89 \\
Dolly 5k & 62.11 & 43.54 & 42.30 & 33.73 & 59.00 & 34.57 & 45.88 \\
\midrule
Dolly 5k + ThaiInstruct 5k & 58.22 & 38.84 & 36.64 & 35.75 & 63.00 & 33.35 & 44.30 \\
Dolly 10k & 63.78 & 44.42 & 40.35 & 34.22 & 64.00 & 32.67 & 46.57 \\
\midrule
Dolly 7.5k + ThaiInstruct 7.5k & 61.11 & 40.45 & 38.41 & 32.38 & 66.20 & 33.33 & 45.31 \\
Dolly 15k & 63.22 & 43.68 & 38.58 & 34.89 & 62.60 & 33.83 & 46.14 \\
\bottomrule
\end{tabular}}
\caption{NLU (Accuracy) benchmark results}
\label{tab:dataset-ablation-nlu}
\end{table*}

\begin{table*}[ht]
\centering
\small
\resizebox{\textwidth}{!}{%
\begin{tabular}{l|cccccc|c}
\toprule
\textbf{Model} & \textbf{Belebele} & \textbf{M3 Exam} & \textbf{Thai Exam} & \textbf{Wisesight} & \textbf{xCopa} & \textbf{xNLI} & \textbf{Average} \\
\midrule
\multicolumn{8}{l}{\textbf{Llama-3.1-8B}} \\
Alpaca 52k + ThaiInstruct 28k & 65.56 & 39.53 & 36.99 & 56.31 & 69.20 & 33.33 & 50.15 \\
Alpaca 52k & 65.67 & 39.21 & 33.81 & 49.42 & 70.00 & 32.18 & 48.48\\
\midrule
Dolly 15k + ThaiInstruct 28k & 62.89 & 39.99 & 35.40 & 34.97 & 64.80 & 33.31 & 45.56 \\
Dolly 15k & 60.67 & 35.93 & 31.50 & 34.71 & 60.00 & 32.00 & 42.47 \\
\midrule
\multicolumn{8}{l}{\textbf{Gemma-2-9B}} \\
Alpaca 52k + ThaiInstruct 28k & 78.33 & 50.78 & 46.73 & 33.96 & 74.40 & 35.49 & 53.95 \\
Alpaca 52k & 79.33 & 49.49 & 47.08 & 36.17 & 72.00 & 35.31 & 53.23 \\
\midrule
Dolly 15k + ThaiInstruct 28k & 79.00 & 50.78 & 47.96 & 28.94 & 78.60 & 34.01 & 53.88 \\
Dolly 15k & 75.11 & 47.32 & 43.36 & 33.10 & 71.60 & 38.10 & 51.43 \\
\midrule
\multicolumn{8}{l}{\textbf{SEA-LIONv2-8B}} \\
Alpaca 52k + ThaiInstruct 28k & 59.44 & 40.18 & 34.51 & 34.03 & 60.00 & 33.07 & 43.87 \\
Alpaca 52k & 57.00 & 40.77 & 34.87 & 39.20 & 58.60 & 33.21 & 43.94\\
\midrule
Dolly 15k + ThaiInstruct 28k & 57.22 & 39.16 & 36.11 & 36.54 & 61.20 & 33.33 & 43.93 \\
Dolly 15k & 63.22 & 43.68 & 38.58 & 34.89 & 62.60 & 33.83 & 46.14 \\
\bottomrule
\end{tabular}}
\caption{Full datasets: NLU results}
\label{tab:full-dataset-nlu}
\end{table*}

\begin{table*}[ht]
\centering
\resizebox{\textwidth}{!}{%
\begin{tabular}{l|ccccc}
\toprule
\textbf{Model} & \textbf{Flores200 Eng-Th (BLEU)} & \textbf{Flores200 Th-Eng (BLEU)} & \textbf{XLsum (RougeL)} & \textbf{iApp QA (RougeL)} \\
\midrule
\multicolumn{5}{l}{\textbf{Llama-3.1-8B}} \\
Alpaca 5k + ThaiInstruct 5k & 2.00 & 4.25 & 13.29 & 3.90 \\
Alpaca 10k & 2.59 & 5.58 & 14.21 & 7.89 \\
\midrule
Alpaca 10k + ThaiInstruct 10k & 1.86 & 3.00 & 13.27 & 3.82 \\
Alpaca 20k & 1.82 & 3.75 & 13.34 & 4.94 \\
\midrule
Alpaca 15k + ThaiInstruct 15k & 1.35 & 2.37 & 13.37 & 3.80 \\
Alpaca 30k & 2.95 & 3.98 & 13.49 & 5.81 \\
\midrule
Dolly 2.5k + ThaiInstruct 2.5k & 2.20 & 5.30 & 13.30 & 4.13 \\
Dolly 5k & 2.83 & 1.13 & 12.49 & 6.60 \\
\midrule
Dolly 5k + ThaiInstruct 5k & 2.08 & 0.63 & 13.24 & 3.86 \\
Dolly 10k & 1.99 & 0.71 & 12.31 & 5.61 \\
\midrule
Dolly 7.5k + ThaiInstruct 7.5k & 1.94 & 1.01 & 13.31 & 3.87 \\
Dolly 15k & 2.23 & 0.96 & 12.48 & 3.71 \\
\midrule
\multicolumn{5}{l}{\textbf{Gemma-2-9B}} \\
Alpaca 5k + ThaiInstruct 5k & 2.16 & 2.33 & 12.63 & 3.65 \\
Alpaca 10k & 1.64 & 1.13 & 10.45 & 3.23 \\
\midrule
Alpaca 10k + ThaiInstruct 10k & 2.13 & 1.90 & 12.50 & 3.67 \\
Alpaca 20k & 1.05 & 1.84 & 10.84 & 3.05 \\
\midrule
Alpaca 15k + ThaiInstruct 15k & 2.12 & 1.83 & 12.38 & 3.65 \\
Alpaca 30k & 0.96 & 1.24 & 8.55 & 2.90 \\
\midrule
Dolly 2.5k + ThaiInstruct 2.5k & 2.38 & 1.12 & 12.51 & 3.63 \\
Dolly 5k & 2.30 & 0.48 & 11.90 & 3.25 \\
\midrule
Dolly 5k + ThaiInstruct 5k & 2.41 & 0.67 & 12.57 & 3.66 \\
Dolly 10k & 2.46 & 0.57 & 11.97 & 3.19 \\
\midrule
Dolly 7.5k + ThaiInstruct 7.5k & 2.53 & 0.73 & 12.54 & 3.70 \\
Dolly 15k & 2.40 & 0.56 & 12.28 & 3.23 \\
\midrule
\multicolumn{5}{l}{\textbf{SEA-LIONv2-8B}} \\
Alpaca 5k + ThaiInstruct 5k & 25.91 & 43.03 & 14.73 & 24.05 \\
Alpaca 10k & 21.46 & 34.56 & 18.02 & 32.68 \\
\midrule
Alpaca 10k + ThaiInstruct 10k & 24.72 & 23.28 & 15.06 & 20.03 \\
Alpaca 20k & 20.83 & 27.61 & 18.77 & 32.87 \\
\midrule
Alpaca 15k + ThaiInstruct 15k & 25.74 & 15.41 & 15.31 & 17.31 \\
Alpaca 30k & 20.57 & 10.58 & 18.87 & 32.48 \\
\midrule
Dolly 2.5k + ThaiInstruct 2.5k & 26.20 & 46.99 & 15.99 & 37.65 \\
Dolly 5k & 25.90 & 25.15 & 19.58 & 51.73 \\
\midrule
Dolly 5k + ThaiInstruct 5k & 25.19 & 6.09 & 15.05 & 32.38 \\
Dolly 10k & 25.24 & 3.56 & 18.83 & 55.86 \\
\midrule
Dolly 7.5k + ThaiInstruct 7.5k & 24.60 & 2.48 & 14.94 & 30.00 \\
Dolly 15k & 25.31 & 3.30 & 18.65 & 52.08 \\
\bottomrule
\end{tabular}}
\caption{Dataset ablation: NLG results}
\label{tab:dataset-ablation-nlg}
\end{table*}

\begin{table*}[ht]
\centering
\small
\resizebox{\textwidth}{!}{%
\begin{tabular}{l|ccccc}
\toprule
\textbf{Model} & \textbf{Flores200 Eng-Th (BLEU)} & \textbf{Flores200 Th-Eng (BLEU)} & \textbf{XLsum (RougeL)} & \textbf{iApp QA (RougeL)} \\
\midrule
\multicolumn{5}{l}{\textbf{Llama-3.1-8B}} \\
Alpaca 52k + ThaiInstruct 28k & 2.01 & 2.32 & 13.19 & 3.89 \\
Alpaca 52k & 1.91 & 2.51 & 13.18 & 5.83\\
\midrule
Dolly 15k + ThaiInstruct 28k & 1.72 & 0.83 & 13.31 & 3.89\\
Dolly 15k & 2.23 & 0.96 & 12.48 & 3.71 \\
\midrule
\multicolumn{5}{l}{\textbf{Gemma-2-9B}} \\
Alpaca 52k + ThaiInstruct 28k & 2.15 & 1.56 & 12.45 & 3.64 \\
Alpaca 52k & 1.36 & 0.86 & 11.56 & 2.51\\
\midrule
Dolly 15k + ThaiInstruct 28k & 2.32 & 0.61 & 12.46 & 3.65 \\
Dolly 15k & 2.40 & 0.56 & 12.28 & 3.23 \\
\midrule
\multicolumn{5}{l}{\textbf{SEA-LIONv2-8B}} \\
Alpaca 52k + ThaiInstruct 28k & 25.79 & 7.00 & 15.69 & 17.32\\
Alpaca 52k & 25.38 & 3.79 & 18.75 & 32r.70\\
\midrule
Dolly 15k + ThaiInstruct 28k & 24.65 & 2.10 & 14.67 & 17.50\\
Dolly 15k & 25.31 & 3.30 & 18.65 & 52.08 \\
\bottomrule
\end{tabular}}
\caption{Full datasets: NLG results}
\label{tab:full-dataset-nlg}
\end{table*}